\documentclass[journal]{IEEEtran}
\IEEEoverridecommandlockouts
\usepackage{amsmath,amssymb,amsfonts}
\usepackage{graphicx}
\usepackage{textcomp}
\usepackage{float,array}
\usepackage{mwe,tikz}
\usepackage{amsmath}

\usepackage{amsthm}
\usepackage{float}
\usepackage{enumitem}
\usepackage{hyperref}
\usepackage{cite}
\usepackage{rotating}
\usepackage{amssymb}
\usepackage{algpseudocode}
\usepackage{algorithm}
\usepackage{epstopdf}

\usepackage{multirow}
\usepackage[caption=false]{subfig}
\usepackage{dblfloatfix} 

\newcommand{\todo}[1]{{\color{black} #1}}

\DeclareMathOperator*{\argmin}{arg\,min}

\newtheorem{defn}{Definition}

\usepackage{bbm}

\begin{document}

\title{ARREST: A RSSI Based Approach for Mobile Sensing and Tracking of a Moving Object}

\author{Pradipta~Ghosh,~\IEEEmembership{Student Member,~IEEE,}
        Jason~A.~Tran,~\IEEEmembership{Student Member,~IEEE,}
        and~Bhaskar~Krishnamachari,~\IEEEmembership{Member,~IEEE}
\thanks{P. Ghosh, J. Tran, and B. Krishnamachari are with the Department
of Electrical Engineering at the Viterbi School of Engineering, University of Southern California, Los Angeles, CA, 90089
E-mail: {pradiptg, jasontra, bkrishna}@usc.edu.}}

\maketitle

\begin{abstract}

We present Autonomous Rssi based RElative poSitioning and Tracking (ARREST), a new robotic sensing system for tracking and following a moving, RF-emitting object, which we refer to as the Leader, solely based on signal strength information. This kind of system can expand the horizon of autonomous mobile tracking and distributed robotics into many scenarios with limited visibility such as nighttime, dense forests, and cluttered environments. Our proposed tracking agent, which we refer to as the TrackBot, uses a single rotating, off-the-shelf, directional antenna, novel angle and relative speed estimation algorithms, and Kalman filtering to continually estimate the relative position of the Leader with decimeter level accuracy (which is comparable to a state-of-the-art multiple access point based RF-localization system) and the relative speed of the Leader with accuracy on the order of 1 m/s. The TrackBot feeds the relative position and speed estimates into a Linear Quadratic Gaussian (LQG) controller to generate a set of control outputs to control the orientation and the movement of the TrackBot. We perform an extensive set of real world experiments with a full-fledged prototype to demonstrate that the TrackBot is able to stay within 5m of the Leader with: (1) more than $99\%$ probability in line of sight scenarios, and (2) more than $70\%$ probability in no line of sight scenarios, when it moves 1.8X faster than the Leader. For ground truth estimation in real world experiments, we also developed an integrated TDoA based distance and angle estimation system with centimeter level localization accuracy in line of sight scenarios. While providing a first proof of concept, our work opens the door to future research aimed at further improvements of autonomous RF-based tracking.
\end{abstract}


%

\section{Introduction}
Sensing and tracking of a moving object/human by a robot is an important topic of research in the field of robotics and automation for enabling collaborative work environments~\cite{mcclure2009darpa}, including for applications such as fire fighting and exploration of unknown terrains \cite{murphy2004trial, penders2011robot, thrun2004autonomous}. In disaster management, robots can assist by tracking and following first-responders while the team explores an unknown environment~\cite{kumar2004robot}. To achieve this, staying in proximity to the first responders is the key.
Another application context of this field of research is in the Leader--Follower collaborative robotics architecture~\cite{ren2008distributed} where a follower robot is required to track and follow a respective Leader.  
Robotic tracking is also required in smart home environments where robots assist humans in daily activities. In this paper, we focus on this class of tracking problems where the term \textbf{``tracking''} refers to the relative position sensing and control of a robot that is required to stay in proximity to an uncontrolled moving target such as a Leader robot or human.

\textbf{{Our Contribution:}} We propose the Autonomous Rssi based RElative poSitioning and Tracking (ARREST), \emph{a purely Radio Signal Strength Information (RSSI) based single node RF sensing system for joint location, angle and speed estimation and \textbf{bounded distance tracking} of a target moving arbitrarily in 2-D that can be implemented using commodity hardware. } 
In our proposed system, the target, which we refer to as the \textbf{Leader}, carries an RF-emitting device that sends out periodic beacons. The tracking robot, which we refer to as the \textbf{TrackBot}, employs an off-the-shelf directional antenna, novel relative position and speed estimation algorithms, and a Linear Quadratic Gaussian (LQG) controller to measure the RSSI of the beacons and control its maneuvers. {Further, to evaluate the ARREST system in a range of large scale and uncontrolled environments, we developed an integrated Time Difference of Arrival (TDoA) based ground truth estimation system for line of sight (LOS) scenarios that can be easily extended to perform a range of large scale indoor and outdoor robotics experiments, without the need of a costly and permanent  VICON~\cite{vicon} system.}
\begin{figure}[!ht]
 \centering
 \includegraphics[width=0.9\linewidth]{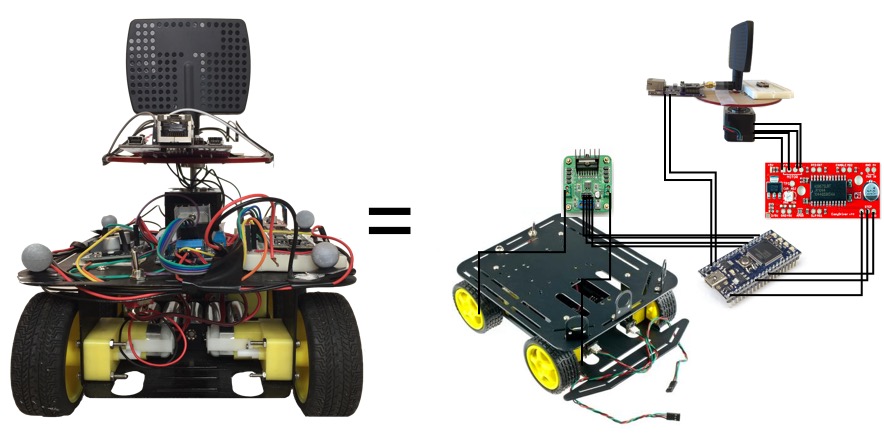}
 \caption{ The TrackBot Prototype}
 \label{fig:robot}
\end{figure}

\textbf{Performance Evaluation Overview:} To analyze and evaluate the ARREST architecture, we develop a hardware prototype (detailed in Section~\ref{sec:hardware}) and perform a set of exhaustive real world experiments as well as emulations. We first perform a set of emulation experiments (detailed in Section~\ref{sec:emulation}) based on real world RSSI data traces collected in various environments. The emulations demonstrate that the TrackBot is able to estimate the target's location with decimeter-scale accuracy, and stay within $5m$ of the Leader (with $\geq 99\%$ probability and with bounded errors in estimations) as long as the Leader's speed is less than or equal to $3m/s$ and the TrackBot's speed is 1.8X times faster than the Leader's speed. 
\emph{Next, using the same parameter setup as in the emulations, we perform a set of small scale real-world tracking experiments (detailed in Section~\ref{sec:real_exp}) in three representative environments: a cluttered indoor room, a long hallway, and a VICON~\cite{vicon} based robotic experiment facility.} {This is followed by a range of large-scale long duration experiments in four representative environments, detailed in Section~\ref{sec:real_exp_large}.} These experiments demonstrate the practicality of our ARREST architecture and validate the emulation results. Moreover, these experiments prove that our ARREST system works well (with $\geq70\%$ probability) in cluttered environments (even in the absence of line of sight) and identify some non-line of sight scenarios where our system can fail.  
{To improve the success rate of our ARREST system in severe non-line of sight (NLOS) situations, we propose a movement randomization technique, detailed in Section~\ref{sec:multipath_adapt}.}
We also compare the ARREST system's performance for varying relative position estimation accuracies offered by different sensing modalities such as camera or infrared in Section~\ref{sec:modality}.

\section{Related Works}
The most popular class of tracking architectures employs vision and laser range finder systems~\cite{papanikolopoulos1993visual,jung2004detecting,Kleinehagenbrock1}. Researchers have proposed a class of efficient sampling and filtering algorithms for vision based tracking such as the Kalman filtering and the particle filtering~\cite{jung2004detecting,schulz2001}. 
There also exist some works that combine vision with range finders~\cite{Kleinehagenbrock1,WendaXu:2015ufa,prassler2001fast}. However, the effectiveness of these sensors crumble when visibility deteriorates or direct line of sight does not exist~\cite{lindstrom2001detecting}.
Moreover, the use of these types of sensors and the processing of their data, namely image processing, increases the form factor and power consumption of the robots which inherently always work under power constraints. 
\emph{In contrast, our proposed RSSI based ARREST system can be developed with low-cost, small form-factor hardware and can be applied in scenarios with limited visibility and non-line-of-sight environments such as cluttered indoor environments and disaster rubble. }


Another class of related works lies within the large body of works in the field of RF Localization in wireless sensor networks~\cite{han2013localization} where robots are employed for localizing static nodes. Graefenstein~\emph{et al.}~\cite{graefenstein2009wireless} employed a rotating antenna on a mobile robot to map the RSSI of a region and exploit the map to localize the static nodes. 
{Similar works have been proposed in the context of locating radio tagged fish or wild animals~\cite{tokekar2011active,tokekar2014multi,vander2014cautious}. 
The works of Zickler and Veloso~\cite{zickler2010rss}, and Oliveira \emph{et al.}~\cite{oliveira2014rssi} on RF-based relative localization are also mentionable. 
In ~\cite{twigg2012rss}, a RSSI based static single radio source localization method is presented by Twigg \emph{et al.} whereas multiple transient static radio source localization problem is discussed in the work of Song, Kim and Yi~\cite{song2012simultaneous}.}
Some researchers have also employed infrared~\cite{pugh2009fast} and ultrasound devices~\cite{rivard2008ultrasonic} for relative localization. One of the most recent significant works on relative localization, which is presented in~\cite{vasisht2016decimeter}, applies a MIMO-based system to localize a single node. {Simulation of a RSSI based constant distance following technique is demonstrated in~\cite{rssi_const_follow} where the leader movement path is predetermined and known to the Follower.  \emph{However, unlike these works, the TrackBot in the ARREST system relies solely on RSSI data not only for the localization of the mobile Leader with unknown movement pattern, but also for autonomous motion control with the goal of maintaining a bounded distance.} The closest state-of-the-art related to our work is presented in~\cite{min2014robotic}. In this work, the authors developed a system that follows the bearing of a directional antenna for effective communication. However, to our knowledge, the maintenance of guaranteed close proximity to the Leader was not discussed in~\cite{min2014robotic}, which is the most important goal in our work. Also, this work employs both RSSI and sonar to determine the orientation of the transmitter antenna. Lastly, compared to their proposed hardware solution which is based on a large robot that carries a laptop as the controller, our solution is low power, small size, and requires much less processing power.}

On LQG related works, Bertsekas~\cite{bertsekas1995dynamic} has demonstrated that a LQG controller can provide the optimal control of a robot along a known/pre-calculated path, when the uncertainty in the motion as well as the noise in observations are Gaussian. Extending this concept, Van Den Berg \emph{et al.}~\cite{van2011lqg,van2012lqg} and Tornero \emph{et al.}~\cite{tornero2001multirate} proposed LQG based robotic path planning solutions to deal with uncertainties and imperfect state observations.
\emph{To the best of our knowledge, we are the first to combine RSSI-based relative position, angle, and speed estimation with the LQG controller for localizing and tracking a moving RF-emitting Object. }

\section{Problem Formulation} 
\label{sec:prob_form} 
In this section, we present
the details of our tracking problem and our mathematical formulation based on
both a 2D  global frame of reference, $\mathcal{R}_{G}$, and the TrackBot's 2D
local frame of reference at time $t$, $\mathcal{R}_{F}(t)$.
Let the location of
the \emph{Leader} at time $t$ be represented as {\small
$\mathbf{X}_{L}(t)=(x_L(t),y_L(t))$} in $\mathcal{R}_{G}$.
The \emph{Leader} follows an
unknown path, $\mathcal{P}_L$.  
Similarly,
let the position of the TrackBot at any time instant $t$ be denoted by {\small
$\mathbf{X}_F(t)=(x_F(t),y_F(t))$}. The maximum speeds of the Leader and the TrackBot
are $v_L^{max}$ and $v_F^{max}$, respectively. 
For simplicity, we discretize the
time with steps of $\delta t > 0$ and use the notation $n$ to refer to the
$n^{th}$ time step i.e., $t = n \cdot \delta t$.
Let {\small $d[n]=||\mathbf{X}_L[n]-\mathbf{X}_F[n]||_2$} be the distance
between the TrackBot and the Leader at time-slot $n$, where $||.||_2$ denotes the
$L_2$ norm. Then, with $D_{th}$ denoting the max distance allowed between the
Leader (L) and the TrackBot (F), the objective of tracking is to plan the
TrackBot's path, $\mathcal{P}_F$, such that {\small$\mathbb{P} \left(d[n] \leq
D_{th}\right) \approx 1 \ \  \forall n$} where $\mathbb{P}(.)$ denotes the
probability. 

\begin{figure}[!ht]
 \centering
 \includegraphics[width=0.7\linewidth]{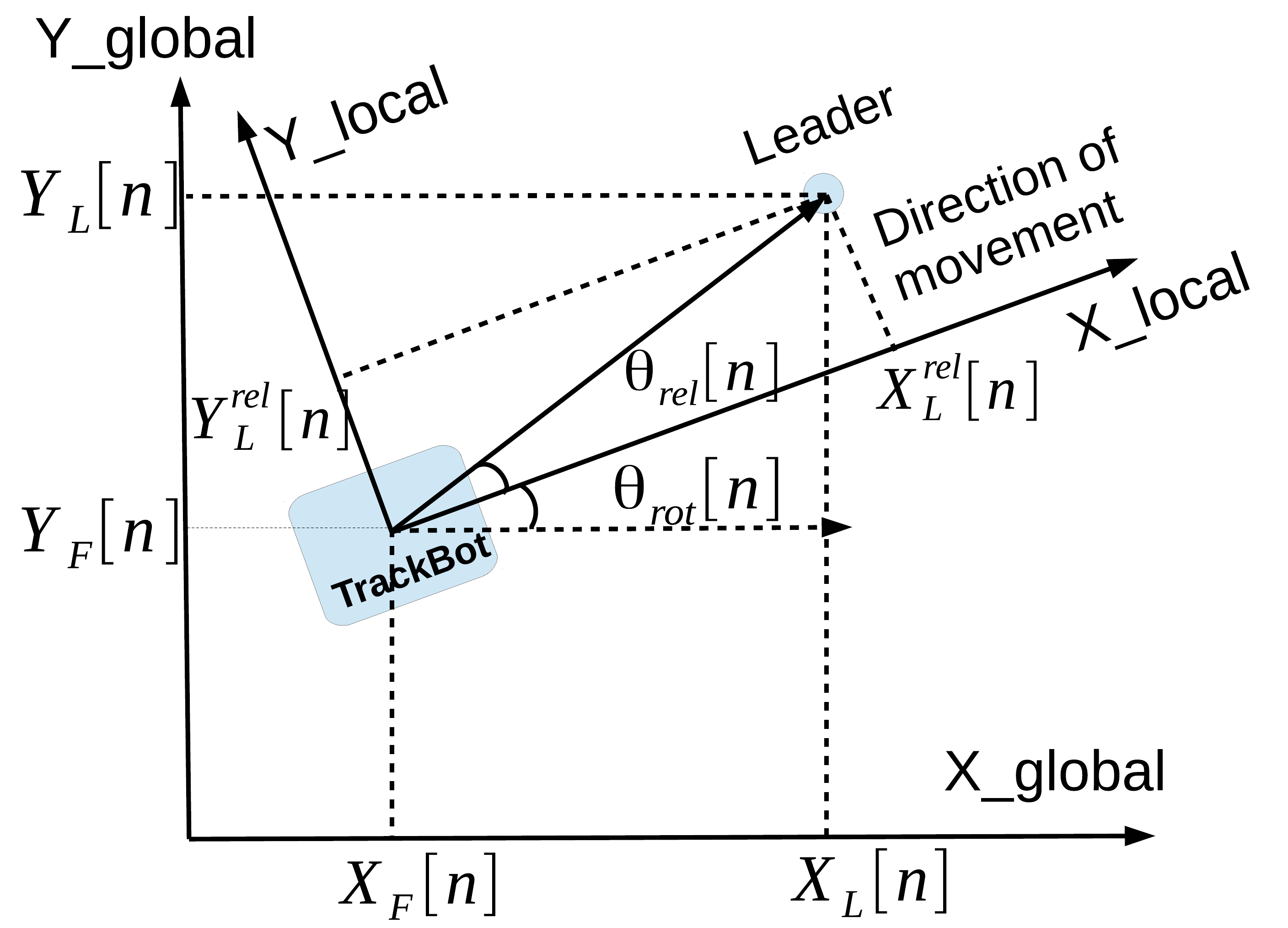}
 \caption{Coordinate System Illustration}
 \label{fig:co_ordinate}
\end{figure}

Realistic deployment scenarios typically do not have a global frame of
reference.  Thus, we formulate a local frame of reference, $\mathcal{R}_{F}[n]$,
with the origin representing the location of the TrackBot, {\small
$\mathbf{X}_F[n]$}.  Let the robot's forward and backward movements at any time
instant $n$ be aligned with the X-axis of $\mathcal{R}_{F}[n]$.  Also, let the
direction perpendicular to the robot's forward and backward movements be aligned
with the Y-axis of $\mathcal{R}_{F}[n]$.  This local frame of reference is
illustrated in Fig.~\ref{fig:co_ordinate}. Note that in our real system all
measurements by the TrackBot are in $\mathcal{R}_{F}[n]$. In order to convert
the position of the Leader in $\mathcal{R}_{F}[n]$ from $\mathcal{R}_{G}$ or
vice versa for simulations and emulations, we need to apply coordinate
transformations.  Let the relative angular orientation of $\mathcal{R}_{F}[n]$
with respect to $\mathcal{R}_{G}$ be $\theta_{rot}[n]$ and the position of the
Leader in $\mathcal{R}_{F}[n]$ be {\small
$\mathbf{X}_{L}^{rel}[n]=(x_L^{rel}[n],y_L^{rel}[n])$}. Then:

{  \small
\begin{equation}
     \begin{bmatrix}
         x_L[n] \\
         y_L[n]\\
         1 \\
        \end{bmatrix}
        =
        \begin{bmatrix}
         \cos (\theta_{rot}[n]) & -\sin (\theta_{rot}[n]) & x_F[n]\\
         \sin (\theta_{rot}[n]) & \cos (\theta_{rot}[n])  & y_F[n]\\
         0 & 0 & 1 \\
        \end{bmatrix}
        \begin{bmatrix}
         x_L^{rel}[n] \\
         y_L^{rel}[n]\\
         1 \\
        \end{bmatrix}
\end{equation}
}

\noindent and  {$\theta_{rel}[n]=\arctan (y_L^{rel}[n]/x_L^{rel}[n])$} is the Leader's direction in {\small$\mathcal{R}_{F}[n]$}.
To restate the objective of tracking in terms of the local coordinates,
{$\mathbb{P} \left(d[n] \leq D_{th}\right) \approx 1 \ \  \forall t$} where { $d[n]=||\mathbf{X}_{L}^{rel}[n]||_2 = ({x_L^{rel}[n]}^2 + {y_L^{rel}[n]}^2)^{1/2}$.}


\section{The ARREST System}

\begin{figure}[!ht]
 \centering
 \includegraphics[width=0.9\linewidth]{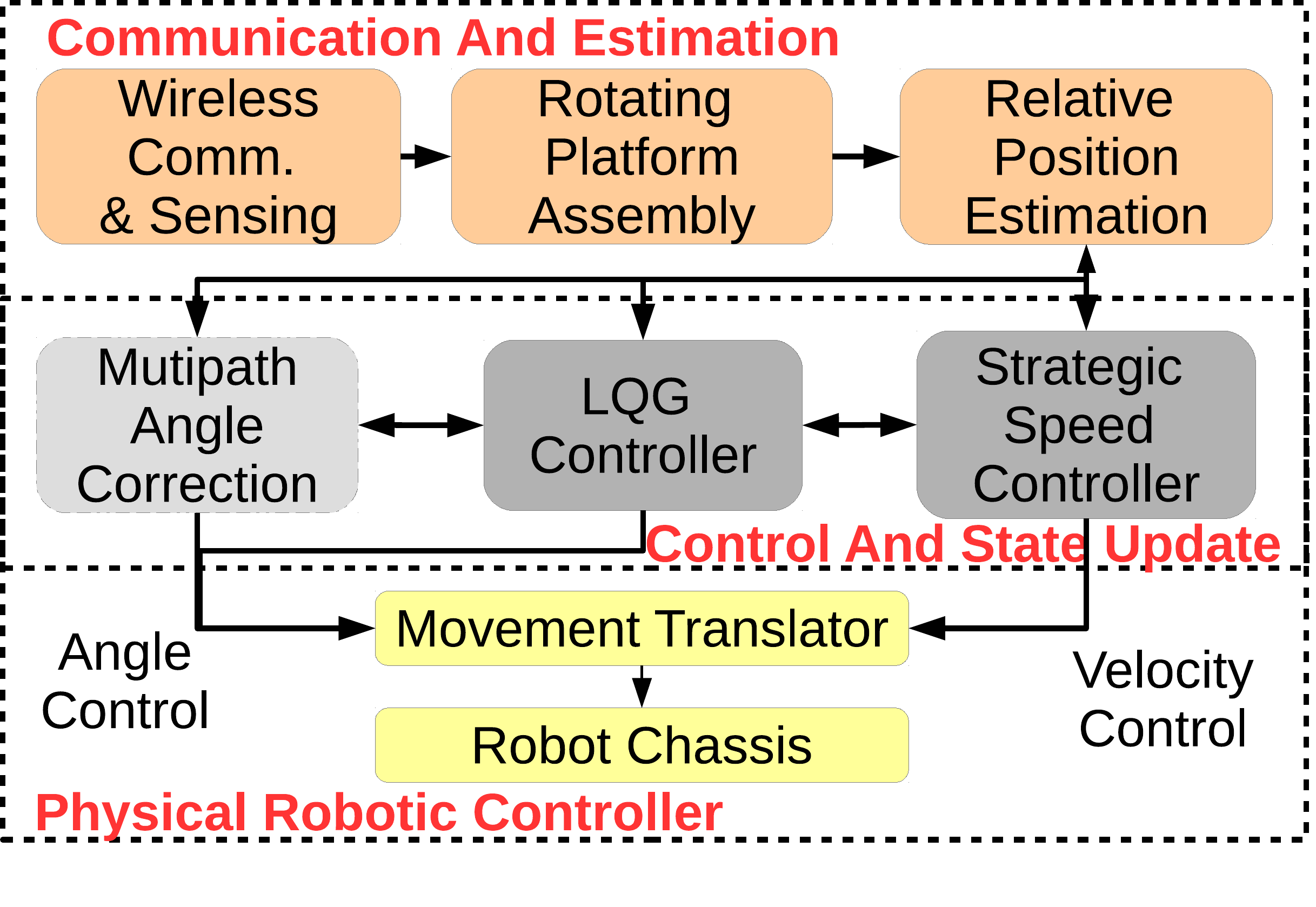}
 \caption{ The ARREST Architecture}
 \label{fig:arrest}
\end{figure}

In this section, we discuss our proposed system solution for RSSI based relative position sensing and tracking. 
In the ARREST system, the Leader is a robot or a human carrying a device that periodically transmits RF beacons, and the TrackBot is a robot carrying a directional, off-the-shelf RF receiver. 
As shown in Fig.~\ref{fig:arrest}, the ARREST architecture consists of three layers: Communication ANd Estimation (CANE), Control And STate update (CAST), and  Physical RobotIc ControllEr (PRICE).
In order to track the Leader, the TrackBot needs sufficiently accurate estimations of both the Leader's relative position ($\mathbf{X}_{L}^{rel}$) and relative speed ($v_{rel}$). Thus, at any time instant $[n]$, we define the state of the TrackBot as a 3-tuple: 
\textbf{{\small $\mathbf{\mathcal{S}}[n] = \begin{bmatrix} d^e[n] ,&    v^{e}_{rel}[n] ,& \theta_{rel}^e[n] \end{bmatrix}$} where the superscript $e$ refers to the estimated values,} $d^e[n]=||\mathbf{X}_{L}^{rel}[n]||_{2}$ refers to the estimated distance at time $n$, $v^{e}_{rel}[n]$ refers to the relative speed of the TrackBot along the X-axis of $\mathcal{R}_F[n]$ with respect to the Leader, and $\theta_{rel}^e[n]$ refers to the angular orientation (in radians) of the Leader in $\mathcal{R}_F[n]$. 

\textbf{CANE:} The function of the CANE layer is to measure RSSI values from the beacons and approximate the Leader's position relative to the TrackBot, (i.e., $d^e[n]$ and $\theta_{rel}^e[n]$). 
The CANE layer is broken down into three modules: Wireless Communication and Sensing, Rotating Platform Assembly, and Relative Position Estimation. 
At the beginning of each time slot $n$, the Wireless Communication and Sensing module and the Rotating Platform Assembly perform a $360^\circ$ RSSI sweep by physically rotating the directional antenna while storing RSSI measurements of successful beacon receptions into the vector $\mathbf{r_v}[n]$.
The Relative Position Estimation module uses $\mathbf{r_v}[n]$ to approximate the relative position of the Leader by leveraging pre-estimated directional gains of the antenna, detailed in Section~\ref{Sec:estimation}. 

%

\textbf{CAST:} The functions of the CAST layer is to maintain the 3-tuple state estimates and to generate control commands based on current and past observations to send to the PRICE layer. {The CAST layer consists of two different modules: the Linear Quadratic Gaussian (LQG) Controller and the Strategic Speed Controller. We also have a special, case specific module called Multipath Angle Correction for severely cluttered environments (explained further in Section~\ref{sec:multipath_adapt}).} The Strategic Speed Controller estimates the relative speed of the Leader by exploiting past and current state information and generates the speed control signal in conjunction with the LQG controller. The term \textbf{``Strategic''} is used to emphasize that we propose two different strategies, Optimistic and Pragmatic, for the relative speed approximation as well as speed control of the TrackBot (detailed in Section~\ref{sec:vel}). The LQG controller incorporates past state information, past control information,
and relative position and speed approximations to: (1) generate the system's instantaneous state, (2) determine how much to rotate the TrackBot itself, and (3) determine what should be the TrackBot's relative speed. The state information generated by the LQG controller is directly sent to the Strategic Speed Controller to calculate the absolute speed of the TrackBot.
The details of our LQG controller formulation are discussed in Section~\ref{sec:lqg}.

\textbf{PRICE:} The goal of the PRICE layer is to convert the control signals from the CAST layer into actual translational and rotational motions of the TrackBot. 
It consists of two modules: Movement Translator and Robot Chassis. The Movement Translator maps the control signals from the CAST layer to a series of platform-specific Robot Chassis motor control signals (detailed in Section~\ref{sec:hardware}). 

\subsection{Proposed LQG Formulation}
\label{sec:lqg}
In our proposed solution, we first formulate the movement control problem of the TrackBot as a discrete time Linear Quadratic Gaussian (LQG) control problem. A LQG controller is a combination of a Kalman Filter with a Linear Quadratic Regulator (LQR) that is proven to be the \textbf{optimal controller} for linear systems with Additive White Gaussian Noise (AWGN) and incomplete state information~\cite{athans1971role}. 
The linear system equations for any discrete LQG problem can be written as:

{\small
   \begin{equation}
    \begin{split}
        &\mathbf{\mathcal{S}}[n+1]=A_n \mathbf{\mathcal{S}}[n]+B_n \mathbf{U}[n] +\mathbf{Z}[n] \\  
        &\mathbf{O}[n]=C_n \mathbf{\mathcal{S}}[n]+\mathbf{W}[n]
    \end{split}
\end{equation}
}

\noindent where $A_n$ and $B_n$ are the state transition matrices, $\mathbf{U}[n]$ is the LQG control vector, $\mathbf{Z}[n]$ is the system noise, $\mathbf{O}[n]$ is the LQG system's observation vector, $C_n$ is the state-to-observation transformation matrix, and $\mathbf{W}[n]$ is the observation noise at time $n$.
A LQG controller first predicts the next state based on the current state and the signals generated by the LQR. Next, it applies the system observations to update the estimates further and generates the control signals based on the updated state estimates.
\textbf{In our case, {\small $\mathbf{O}[n]=\begin{bmatrix} d^{m}[n], & v_{rel}^{m}[n], & \theta_{rel}^{m}[n] \end{bmatrix}^T$ }(the superscript $m$ refers to measured values).} Moreover, in our case, {the state transition matices $A_n=A$, $B_n=B$, $C_n=C$ are time invariant and the time horizon is infinite as we do not have any control over the Leader's movements.} For a infinite time horizon LQG problem~\cite{bertsekas1995dynamic}, the cost function can be written as:

{\small
   \begin{equation}
   \label{lqg:Cost}
    J=\lim_{N \rightarrow \infty} \frac{1}{N}\mathbb{E}\left(\sum_{n=0}^{N} \mathcal{S}[n]^T \mathbf{Q} \mathcal{S}[n]+ \mathbf{U}[n]^T \mathbf{H} \mathbf{U}[n] \right)
\end{equation}}

\noindent 
where $\mathbf{Q}\geq 0,\mathbf{H}>0$ are the weighting matrices.
The discrete time LQG controller for this optimization problem is:

{\small
   \begin{equation}
    \begin{split}
        &\hat{\mathcal{S}}[n+1]=A\hat{\mathcal{S}}[n]+B\mathbf{U}[n] +K(\mathbf{O}[n+1]-C \{A\hat{\mathcal{S}}[n]+B\mathbf{U}[n]\})\\
        &\mathbf{U}[n]=-L\hat{\mathcal{S}}[n] \qquad \mbox{and} \qquad
        \hat{\mathcal{S}}(0)=\mathbb{E}(\mathcal{S}(0))
    \end{split}
\end{equation}}

\noindent where $\hat{}$ denotes estimates, $K$ is the Kalman gain which can be solved via the algebraic Riccati equation~\cite{lancaster1995algebraic}, and $L$ is the feedback gain matrix.
In our system, the state transition matrix values are as follows:

{\small
   \begin{equation}
     \mathbf{A}=\begin{bmatrix}
         1 & -\delta t & 0 \\
         0 & 1 & 0\\
         0 & 0 & 1 
        \end{bmatrix}
        \mathbf{B}=\begin{bmatrix}
        0 & -\delta t & 0 \\
        0 & 1 & 0 \\
        0 & 0 & -1
        \end{bmatrix}
        \mathbf{C}=\begin{bmatrix}
        1 & 0 & 0 \\
        0 & 1 & 0 \\
        0 & 0 & 1
        \end{bmatrix}
        \label{sec:state}
  \end{equation}
 }
 
\noindent
\emph{where $\delta t$ is the time granularity for the state update. Ideally, within $\delta t$, the TrackBot executes one set of movement control decisions while it also scans RSSI for the next set of control decision (detailed in Sections~\ref{sec:emulation} and~\ref{sec:real_exp}).}
Note that, to solve this optimization problem, we also require the covariance data for the noise, i.e., $ \Sigma_{WW}=\mathbb{E}(\mathbf{W} \mathbf{W}^T),$ and $\ \Sigma_{ZZ}=\mathbb{E}(\mathbf{Z} \mathbf{Z}^T)$. We assume the system noise, $Z[n]$, to be Gaussian and the measurement noise, $W[n]$, to be approximated as Gaussian.

\begin{figure}[!ht]
 \centering
 \includegraphics[width=0.7\linewidth]{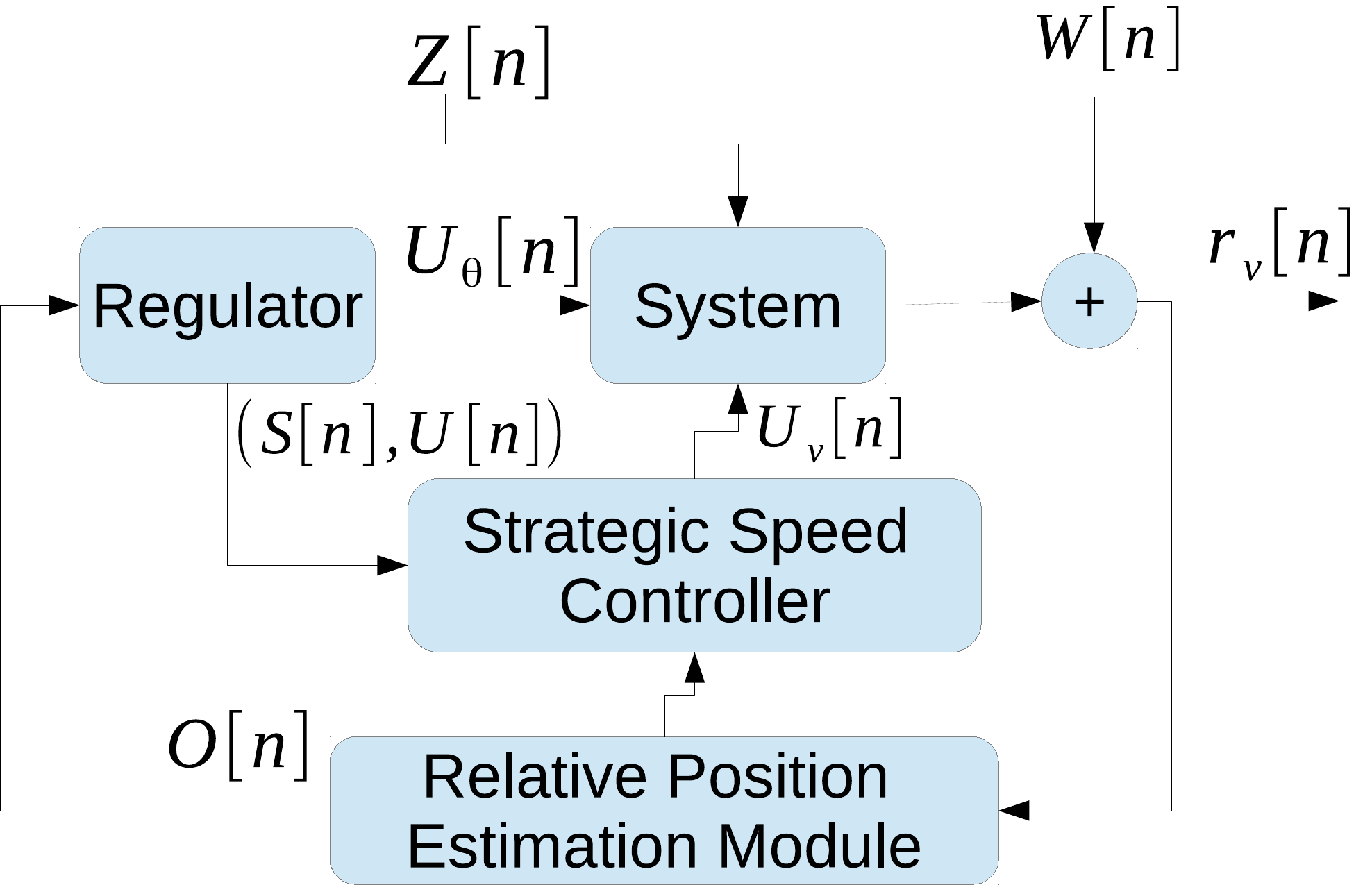}
 \caption{Proposed LQG Controller System}
 \label{fig:system}
\end{figure}
Furthermore, we tweak the LQG controller to send out a rotational control signal after a state update and before generating the LQR control signals, $\mathbf{U}[n]$. The rotational control signal rotates the TrackBot assembly by ${\theta}^{e}_{rel}[n]$ and sets ${\theta}_{rel}^{e}[n]=0$.
This is performed to align the robot toward the estimated direction of the Leader before calculating the movement speed. Thus, we use only the Kalman Filtering part of the LQG controller for angle/orientation control. The reason behind not using the full LQG controller for TrackBot's orientation control lies in the fact that the LQG controller considers sudden rapid change in direction ($\approx 180^\circ$) as a noise and takes a while to correct the course of the TrackBot. More study of this problem is left as a future work. A block diagram of our LQG control system model is presented in Fig.~\ref{fig:system}.


\section{RSSI Based Relative Position and Speed Observations}
\label{Sec:estimation}
In this section, we discuss our methodologies to map the observed RSSI vector, $\mathbf{r_v}[n]$, into the controller observation vector,  $\mathbf{O}[n]$. 

\subsection{Distance Observations}
\label{sec:dist_est}
The RSSI is well known to be a measure of distance if provided with sufficient transceiver statistics such as the transmitter power, the channel path loss exponent, and the fading characteristics.
One of the standard equations for calculating the received power for an omnidirectional antenna is as follows~\cite{rappaport1996wireless}:
{
   \begin{equation}
\label{eqn:power}
\begin{split}
        &P_{r,dBm}=P_{t,dBm}+G_{dB}-\mathcal{L}_{ref}-10\eta \log_{10} \frac{d^m[n]}{d_{ref}} + \psi\\
        &P_{r,dBm}^{ref}=P_{t,dBm}+G_{dB}-\mathcal{L}_{ref} + \psi \\
        &\implies \frac{d^m[n]}{d_{ref}} \approx 10^{\frac{\left(P_{r,dBm}^{ref} -P_{r,dBm} \right) }{10\cdot \eta}} 
       \end{split}
\end{equation}
}

\noindent
where $P_{r,dBm}$ is the received power in dBm, $P_{t,dBm}$ is the transmitter power in dBm, $G_{dB}$ is the gain in dB, $\mathcal{L}_{ref}$ is the path loss at the reference distance $d_{ref}$ in dB, $\eta$ is the path loss exponent, $d^m[n]$ is the distance between the transmitter and receiver, $\psi$ is the random shadowing and multipath fading noise in dB, and $P_{r,dBm}^{ref}$ is the received power at reference distance ($d_{ref}$).
Eqn.~\eqref{eqn:power} is also valid for the average received power for a directional antenna with an average gain of $G_{dB}$. {To calculate the received power for a particular direction $\theta$, we just need to replace $G_{dB}$ in \eqref{eqn:power} with the directional gain of the antenna, $G_{dB}(\theta)$.}
To apply~\eqref{eqn:power} in ARREST, the TrackBot needs to learn the channel parameters such as the $\eta$, $\mathcal{L}_{ref}$, and $d_{ref}$. 
In our proposed system, we assume that the TrackBot has information about the initial distance to the Leader ($d^m(0)$) and the average received power ({\small $P_{r,dBm}^{ref}$}) at reference distance ($d_{ref}$) which we choose to be 1 meter. 
Furthermore, the directional gain, $G_{dB}(\theta)$, and the transmitter power, $P_{t,dBm}$, are known as a part of the system design process. 
Upon initialization of ARREST, the TrackBot performs a RSSI scan by rotating the antenna assembly to generate  $\mathbf{r_v}(0)$ and harnesses the average received power ($P_{r,dBm}$) information to estimate the environment's $\eta$ as follows.
{
   \begin{equation}
    \eta=({P_{r,dBm}^{ref}-P_{r,dBm}})/{10 \log_{10} \frac{d^m(0)}{d_{ref}}}
    \label{eqn:eta_est}
\end{equation}
}

\noindent
Next, the TrackBot applies the estimated $\eta$ and $P_{r,dBm}= avg $ $\{ \mathbf{r_v}[n]\}$ on~\eqref{eqn:power} to map $\mathbf{r_v}[n]$ to the observed distance to the Leader, $d^m[n]$.

\subsection{Angle Observations}
\label{sec:arrest_angle}
One of the main components of our ARREST architecture is the observation of the Angle of Arrival (AoA) of RF beacons solely based on the RSSI data, $\mathbf{r_v}[n]$. There exist three different classes of RF based solutions to determine the AoA. The first class, \textbf{antenna array based approaches}, employs an array of antennas to determine the AoA by leveraging the phase differences among the signals received by the different antennas~\cite{xiong2013arraytrack}. The main difficulty of implementing this class is that very few multi-antenna off-the-shelf radios provide access to phase information. The second class, \textbf{multiple directional antenna based approaches}, employs at least two directional antennas oriented in different directions~\cite{jiang2012alrd} to determine AoA. In this class, the differences among RSSI values from all antennas are utilized to determine the AoA. However, utilizing current off-the-shelf antenna arrays or multiple directional antennas increases the cost, form factor, and complexity of a TrackBot implementation. {We avoid the multiple directional antenna based option also because it requires separate radio drivers for each antenna as well as proper time synchronizations.} Thus, we develop methods contributing to the third class of solutions, which is the use of a single, rotating antenna and the knowledge of the antenna's directional gain pattern to approximate the AoA of RF beacons. 
The core of these methods, called \emph{pattern correlation}, is to correlate the vector of RSSI measurements, $\mathbf{r_v}[n]$, with another vector representing the antenna's known, normalized gain pattern, $\mathbf{g_{abs}}$. At the beginning of each time slot $n$, the TrackBot performs a $360^\circ$ sweep of RSSI measurements to generate the vector, $\mathbf{r_v}[n]$. Then, $\mathbf{r_v}[n]$ is normalized: $\mathbf{g_{m}}=\mathbf{r_v}[n]-\max (\mathbf{r_v}[n])$.  The TrackBot also generates different $\theta$ shifted versions of $\mathbf{g_{abs}}(\theta)$ as follows. 
{
\begin{equation}
\begin{split}
    &\mathbf{r_v}[n]=[ r_{-180},r_{-178.2},\cdots, r_{-1.8},r_{0},r_{1.8},\cdots r_{178.2}]\\
    &\mathbf{g_{m}}=[ r'_{-180},r'_{-178.2},\cdots, r'_{-1.8},r'_{0},r'_{1.8},\cdots r'_{178.2}]\\
    &\mathbf{g_{abs}}(\theta)=[ g_{(-180+\theta)}, \cdots, g_{(0+\theta)}, \cdots,  g_{(178.2+\theta)}]
\end{split}
\label{corr_eqn}
\end{equation} }

\noindent
where $r_{\phi}$ refers to the RSSI measurement, $g_{\phi}$ refers to the antenna gain, and $r'_{\phi}=r_{\phi}-\max \{ \mathbf{r_v}\}$ refers to the observed gain for the antenna orientation of $\phi^\circ$ with respect to the X-axis of $\mathcal{R}_{F}[n]$. The step size of $1.8^\circ$ is chosen based on our hardware implementation's constraints. \textbf{Thus, the possible antenna orientations ($\phi$) are limited to {\small $\Theta=\{-180,\cdots,-1.8,0,$ $\cdots,178.2\}$}.} Next, the TrackBot employs different pattern correlation methods for the AoA observation. Below, we describe three methods in increasing order of complexity. The first method was originally demonstrated by \cite{graefenstein2009wireless}. Through real world experimentation, we develop two additional improved methods.

\subsubsection{Basic Correlation Method} 
\label{sec:aprox1}
The first method of determining AoA correlates $\mathbf{g_{m}}$ with all $\theta$ shifted versions of $\mathbf{g_{abs}}$ and calculates the respective $L_2$ distances. The observed AoA is the $\theta$ at which the $L_2$ distance is the smallest: 
{
\begin{equation}
\begin{split}
    \theta_{rel}^{m} =\argmin_{\theta\in \Theta}  \sum_{k\in \Theta} ||r'_{k}-g_{(k+\theta)}||_2 \cdot \mathbb{I}_{r'_{k}}
\end{split}
\label{corr_eqn1}
\end{equation}
}

\noindent
{\small $\mathbb{I}_{r'_{k}}$} is an indicator function to indicate whether the sample ${r'_{k}}$ exists or not to account for missing samples in real experiments. 

\subsubsection{Clustering Method}
While the first method works well if enough uniformly distributed samples ($\geq 100$ in our implementation) are collected within the $360^\circ$ scan, it fails in scenarios of sparse, non-uniform sampling ($<100$ samples), which occurs in practice due to packet loss due to fading and interference from collocated WiFi devices. In real experiments (mainly indoors), the collected RSSI samples can be uniformly sparse or sometimes batched sparse (samples form clusters with large gaps ($\approx 30^\circ $) between them). 

\begin{defn}
\emph{Angular Cluster:} An angular cluster ($\Lambda$) is a set of valid samples for a contiguous set of angles: {\small $\Lambda=\{k | \mathbb{I}_{r'_{k}}=1 \forall k \in \{\phi_f,\phi_f+1.8,\cdots,\phi_l-1.8, \phi_l\} \}$ } where {\small $\phi_f,\phi_l \in \Theta$} define the boundary of the cluster.
\end{defn}

\noindent
To prevent undue bias from large cardinality clusters that can cause errors in estimating the correlation, we assign a weight ($\omega_{k}$) to each sample ($k$) and use the pattern correlation method as follows.
{
   \begin{equation}
    \theta_{rel}^{m} =\argmin_{\theta \in \Theta} \sum_{k\in \Theta} \omega_{k} \cdot ||r'_{k}-g_{(k+\theta)}||_2 \cdot \mathbb{I}_{r'_{k}}
\label{eqn:new_est}
\end{equation}
}

\noindent
In our weighting scheme, we assign $\omega_{k}=\frac{1}{|\Lambda|}$ where $k \in \Lambda$. Thus, the sum of all weights of the samples from a single cluster sums to $1$, i.e., the weights of the samples are defined by the angular cluster it belongs to.

\subsubsection{Weighted Average Method}
Based on real world experiments, we find that the angle observation based on~\eqref{corr_eqn1}, say $\theta_{m}^1$, gives reasonable error performance if the average cluster size, $\lambda_{a}$, is greater than the average gap size between clusters, $\mu_{a}$. Conversely, the angle observation based on~\eqref{eqn:new_est}, say $\theta_{m}^2$, is better if $\lambda_{a} << \mu_{a}$. 
Thus, as a trade-off between both the basic correlation method and the clustering method, we propose a weighted averaging method described below.
{
\begin{equation}
 \theta_{rel}^{m}=
 \begin{cases}
 \frac{\lambda_{a}}{\mu_{a}}\cdot \theta_{m}^1 + (1-\frac{\lambda_{a}}{\mu_{a}})\cdot \theta_{m}^2 &\mbox{if $\lambda_{a}\leq \mu_{a}$ }\\
 \theta_{m}^1 &\mbox{if $\lambda_{a} > \mu_{a}$ }
 \end{cases}
 \label{eqn:comb_est}
\end{equation}
}

\noindent
\emph{In the rest of the paper, we use the weighted average method for angle observations.}
We compare the performance of all three methods based on real world experiments in Section~\ref{sec:real_est_err}.

\subsection{Speed Observations}
\label{sec:vel}
To fulfill the tracking objective, the TrackBot needs to adapt its speed of movement ($v_F[n]$), according to the Leader's speed ($v_L[n]$). 
In our ARREST architecture, the Strategic Speed Controller uses the relative position observations $(d^m[n],\theta^m_{rel}[n])$ from the CANE layer and the past LQG state estimates to determine the current relative speed, $v_{rel}^{m}[n]$, as well as the Leader's speed, $v_{L}^{m}[n]$. In this context, we employ two different observation strategies. The first strategy, which we refer to as the \emph{Optimistic strategy}, assumes that the Leader will be static for the next time slot and determines the relative speed as follows:
{
   \begin{equation}
\begin{split}
    &v^{m}_{rel}[n]=v^{e}_{rel}[n]-\frac{(d^m[n] -d^e[n]\cdot \cos (\theta_{rel}^m[n]))}{\delta t}\\
    &v_{L}^{e}[n+1]=0    
\end{split}
\end{equation}
}

\noindent
On the other hand, the \emph{Pragmatic Strategy} assumes that the Leader will continue traveling at the observed speed, $v_{L}^{m}[n]$. This strategy determines the relative speed as follows:
{
   \begin{equation}
\begin{split}
    & v_1 =  \{d^m[n]\cdot \cos (\theta_{rel}^m[n]) - d^e[n]\}\\
    & v_2 =  \{d^m[n]\cdot \sin (\theta_{rel}^m[n])\}\\
    &v_L[n]= \frac{\sqrt{v_1^2+v_2^2}}{\delta t} \\
    &\theta_v[n]=\arctan \frac{v_2}{v_1} -\theta_{rel}^m[n] \\
    &v_{L}^{e}[n+1]=v_{L}^{m}[n]=v_L[n] \cdot \cos (\theta_v[n])\\ &v^{m}_{rel}[n]=v_{F}[n]-v_{L}^{m}[n]
\end{split}
\end{equation}
}

\begin{figure}[!ht] 
 \centering
 { \includegraphics[width=\linewidth]{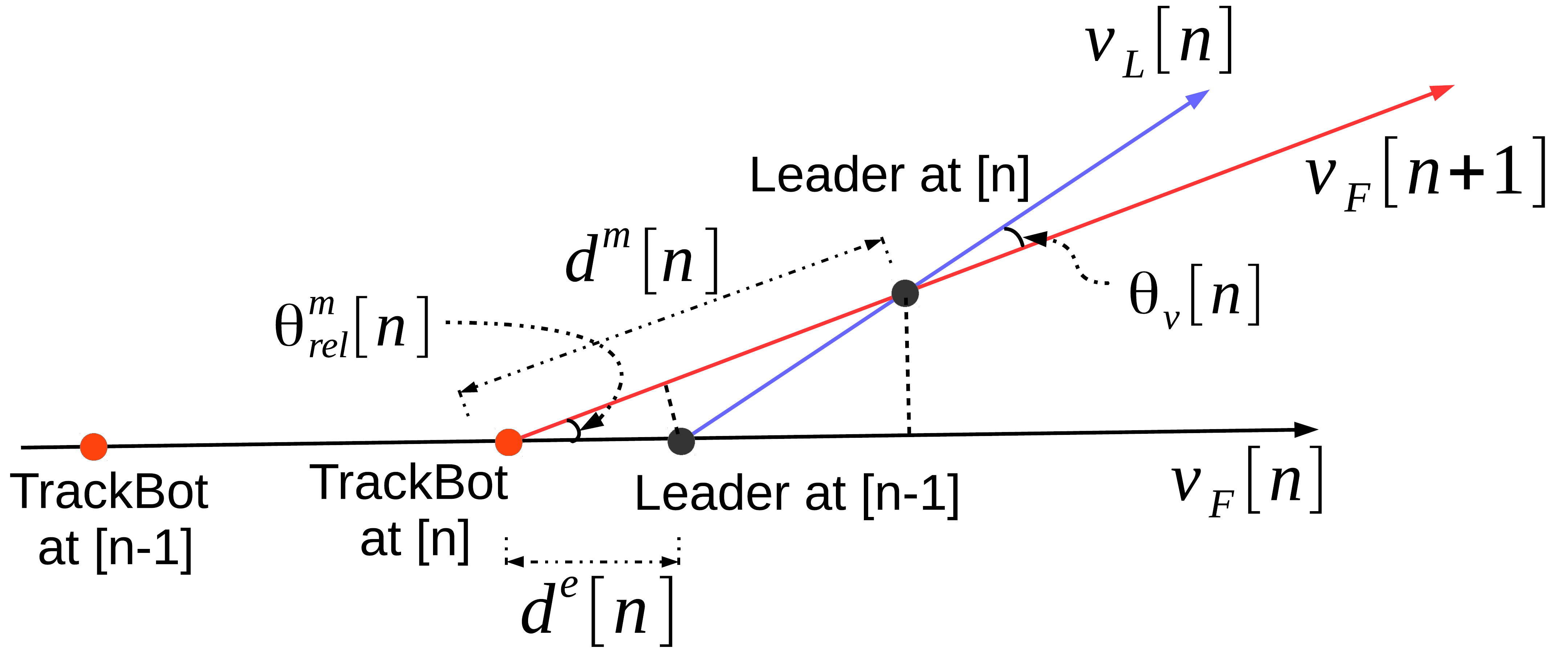}}
 \caption{  Illustration of the Relative Speed Observation}
\label{fig:vel_estimate}
\end{figure}

\noindent
For an illustration of different components of this process, please refer to Fig.~\ref{fig:vel_estimate}. Next, the LQG controller uses the observation vector $\mathbf{O}[n]$ to decide the next state's relative speed, $v^{e}_{rel}[n+1]$ which is used by the Speed Controller to generate the TrackBot's actual speed for next time step, $v_{F}[n+1]=v_{L}^{e}[n+1]+v^{e}_{rel}[n+1]$. 
Note that the speed of the TrackBot, $v_{F}[n]$, is exactly known to itself at any time $n$. 
In addition to the different assumptions about the Leader's speed, the two strategies also differ in how the noise is modeled in the correlation between distance and speed estimations: the Optimistic Strategy assumes that the noise in speed observations are uncorrelated with the noise in distance observations, whereas the Pragmatic strategy assumes strong correlation between distance and speed estimation noise. We compare the performance of both strategies based on emulation and real world experiments in Sections~\ref{sec:comparison_emulation} and~\ref{sec:comparison_real_world}, respectively.



\section{TrackBot Prototype}
\label{sec:hardware}
\subsection{Hardware} We implemented a TrackBot with our ARREST architecture inside a real, low-cost robot prototype presented in Fig.~\ref{fig:robot}. 
For a concise description of our prototype, we list the hardware used for implementation of each of the ARREST components in Table~\ref{tab:hardware_summary}.
{We discuss details of the Time Difference of Arrival (TDOA) based localization system integrated with our ARREST architecture for ground truth estimation separately in Section~\ref{sec:real_exp_large}.}

{\small
\begin{table}[!h]
    \centering
        \caption{ARREST Hardware Implementation}
    \resizebox{\linewidth}{!}
    {\small
    \begin{tabular}{|m{0.025\linewidth}|m{0.275\linewidth}|m{0.7\linewidth}|}
    \hline
         & Module & Hardware\\ \hline
          & Wireless Communication  and  Sensing & OpenMote\cite{Openmote}; Rosewill Directional Antenna (Model RNX-AD7D) \\ \cline{2-3}
        \rotatebox{90}{\multirow{2}{*}{\colorbox{blue!30}{CANE}}}    & Rotating Platform Assembly &   Nema 17 (4-wire bipolar Stepper Motor); EasyDriver - Stepper Motor Driver; mbed NXP LPC1768~\cite{mbed_ref}\\ \cline{2-3}
            & Relative Position Estimation & mbed NXP LPC1768~\cite{mbed_ref}\\ \hline
         \multicolumn{2}{|l|}{\colorbox{blue!30}{CAST}}  &  mbed NXP LPC1768~\cite{mbed_ref}\\ \hline
         \parbox[t]{2mm}{\multirow{2}{*}{\rotatebox[origin=c]{90}{\colorbox{blue!30}{PRICE}}}} & Movement Translator & mbed NXP LPC1768~\cite{mbed_ref} \\ \cline{2-3}
          & Robot Chassis &  Baron-4WD Mobile Platform, L298N Stepper Motor Driver Controller Board, HC-SR04 Ultrasonic Sensor~\cite{ultra_1}\\ \hline \hline
    \end{tabular}}
    
    \resizebox{\linewidth}{!}
    {\small
    \begin{tabular}{|m{0.2\linewidth}|m{0.8\linewidth}|}
    \hline
     OpenMote~\cite{Openmote} &  TI 32-bit CC2538 @ 32 MHz with 512KB Flash memory, 32KB RAM, 2.4GHz IEEE 802.15.4-based Transceiver connected via SMA plug \\  \hline
     mbed NXP-LPC1768~\cite{mbed_ref} $\mu$-processor &  32-bit ARM Cortex-M3 core @ 96MHz, 512KB FLASH, 32KB RAM; \textbf{Interfaces:} built-in Ethernet, USB Host and Device, CAN, SPI, I2C, ADC, DAC, PWM and other I/O interfaces \\  \hline
     Rosewill RNX-AD7D Directional Antenna  & \textbf{Mode 1:} \textbf{Frequency:} 2.4GHz, \textbf{Max Gain:} 5dBi, \textbf{HPBW:} $70^\circ$ \textbf{Mode 2:} \textbf{Frequency:} 5GHz, \textbf{Max Gain:} 7dBi, \textbf{HPBW:} $50^\circ$\\ \hline
     Nema 17 Stepper Motor & \textbf{Dimension:} 1.65"x1.65"x1.57", \textbf{Step size:} 1.8 degrees (200 steps/rev), \textbf{Rated current:} 2A, \textbf{Rated resistance:} 1.1 Ohms\\ \hline
     HC-SR04~\cite{ultra_1} & \textbf{Operating Voltage:} 5V DC, \textbf{Operating Current:} 15mA, \textbf{Measure Angle:} $15^\circ$, \textbf{Ranging Distance:} 2cm - 4m\\ \hline
    \end{tabular}}
    \label{tab:hardware_summary}
\end{table}
}

In the TrackBot prototype, the directional antenna and the OpenMote are mounted on top of a stepper motor using a plate. 
While we use two microprocessors (the OpenMote and the mbed), the system can be implemented using one microprocessor. We choose to use two in this prototype to work around wiring issues and work around the lack of sufficient GPIO pins on the OpenMote.
The OpenMote is only used for RF sensing while the mbed is used to implement the rest of the ARREST modules. 
For programming of the mbed,  we use the mbed Real Time Operating System~\cite{rtos}. 
The mbed sends control signals to the stepper motor to rotate it in precise steps of $1.8^\circ$. \emph{Each consecutive $360^\circ$ antenna rotations alternate between clockwise and anti-clockwise because this: (1) prevents any wire twisting between the mbed and OpenMote and (2) compensates for the stepper motor's movement errors.} 
The mbed communicates with other H/W components via GPIO pins and High Level Data Link Control (HDLC) Protocol~\cite{gelenbe1978performance} based reliable serial line communication.

In the current prototype, the maximum speed
of the robot is $30cm/s$.
Due to synchronization issues on the mbed when trying to simultaneously rotate the antenna and move the robot chassis, the antenna assembly sometimes does not return to its initial position after a complete rotation. 
To solve this issue while avoiding complex solutions (e.g., via a feedback-based offset control mechanism), the TrackBot instead first performs a RSSI scan and then moves the chassis.
Ideally, the antenna can rotate $360^\circ$ in $1 s$ while collecting $200$ samples.
However, we choose to slow the scan down to a duration of $2 s$ to cope with the occasional occurrence of sparse RSSI samples. 
Moreover, to keep the movement simple, the TrackBot first rotates to the desired direction and then moves straight with the desired speed. 
The wheels of the robot are controlled using PWM signals from the mbed with a period of $2 s$. We choose a $2 s$ period for robot rotation as one $2 s$ pulse width equates to a chassis rotation amount of $\approx 180^\circ$. We also choose the same period length ($2 s$) for forward movement which caps the speed of the robot at $60/6 =10 cm/s$ (including $2 s$  of RSSI scan). 
The whole system is powered by five AA batteries which can run for a total of $\approx 3-4$ hours.
{We also implemented a very simple obstacle avoidance mechanism by employing a single HC-SR04 range finder in the front bumper of the chassis and protection bumpers on the other sides. While moving forward, if the ultrasound detects an object at a distance less than $10$cm, it stops the TrackBot's movement immediately.}

The Leader node is currently implemented as an OpenMote transmitting beacons with the standard omnidirectional antenna and a transmit power of $7dBm$. For programming of the OpenMotes, {we use the RIOT operating systems~\cite{riot,baccelli2013riot}. The Leader implementation is capable of transmitting $200$ packets/second.}

\subsection{ARREST System Parameter Setup}
\label{sec:lqg_param}
We discuss here the choices of the LQG Controller parameters such as the $\mathbf{Q}$, $\mathbf{H}$, $\Sigma_{WW}$ and $\Sigma_{ZZ}$.

\subsubsection{Cost Parameters Setup} In the cost function of LQG, the matrix
$\mathbf{Q}$ determines the weights of different states on the overall cost,
$J$. In our case, $\mathbf{Q}$ is a $3\times 3$ positive definite matrix with
nonzero diagonal terms: 

{\small
  \begin{equation}
\mathbf{Q}=\begin{bmatrix}
        Q_d & 0 & 0 \\
        0 & Q_v & 0 \\
        0 & 0 & Q_\theta
        \end{bmatrix}
\end{equation}
}

\noindent Our main goal is to keep the distance as well as the relative angle to
be as low as possible while keeping emphasis on the distance. From this
perspective, the weights in increasing order should be $Q_v$, $Q_\theta$ and
$Q_d$, respectively. Furthermore, focusing on one particular aspect such as the
distance has detrimental effects on the other aspects. Thus, we perform a set of
experiments to find a good trade-off between $Q_v$, $Q_\theta$ and $Q_d$ where
we vary one parameter while keeping the rest of them fixed.
For example, we vary the
value of $Q_d$ by keeping $Q_v$ and $Q_\theta$ fixed. Based on these
experiments, we opt for the following settings: $Q_v=0.1$, $Q_\theta=1$ and
$Q_d=10\cdot v_{L}^{max}$ where $v_{L}^{max}$ is the maximum speed of the
Leader. With these settings, our system performs better than any other explored
settings. Furthermore, $\mathbf{H}$ is chosen to be a $3\times 3$ Identity
matrix. Note that, the values of $\mathbf{Q}$ and $\mathbf{H}$ are strategy
(Optimistic or Pragmatic) independent.

\subsubsection{Noise Covariance Matrix Parameters Setup}
The noise covariance matrices, $\Sigma_{WW}$ and $\Sigma_{ZZ}$, need to be properly set for a good state estimation in the presence of noise and imperfect/partial state observations. The system noises are assumed to be i.i.d normal random variables with $\Sigma_{ZZ}$ being a $3\times 3$ identity matrix. On the other hand, the observation noise covariance matrix requires separate settings for the different strategies. For the Optimistic strategy, we assume that the observation noises are uncorrelated, whereas, for the Pragmatic strategy, the distance estimation errors and the relative speed estimation errors are highly correlated with variances proportional to $v_{L}^{max}$. A set of empirically determined values of $\Sigma_{WW}$ for the Optimistic and the Pragmatic strategies are as follows.

{ \small 
\begin{equation}
\Sigma_{WW}^{Op}=
\begin{bmatrix} 
4 & 0 & 0 \\ 
0 & 2 & 0 \\
0 & 0 & 1 
\end{bmatrix}
,
\Sigma_{WW}^{Pg}=
\begin{bmatrix} 
1 & v_{L}^{max} & 0 \\ 
v_{L}^{max} & (v_{F}^{max})^2 & 0 \\ 
0 & 0 & 0.1 
\end{bmatrix}
\end{equation}}

\noindent
where $Op$ and $Pg$ refers to the Optimistic and the Pragmatic strategies, respectively.

\section{Baseline Analysis via Emulation}
\label{sec:emulation}
In this section, we perform a thorough evaluation and setup the different parameters such as the LQG covariance matrix (discussed in Section~\ref{sec:lqg_param}) of the ARREST architecture via a set of emulation experiments. 
We use the emulation experiment results as a baseline for our real-world experiments.

\subsection{Method}
{
We employ our hardware prototypes, discussed in Section~\ref{sec:hardware}, to collect sets of RSSI data in cluttered indoor and outdoor environments for a set of representative distances, $\mathcal{D}$, and angles $\Theta$.
Next, we use the collected samples to interpolate the RSSI samples for any random configuration $\mathcal{C}=(d,\theta_{rel})$, where $d \in \mathbb{R}^+$ and $\theta_{rel} \in [-180, 180)$, as follows: $ r^e=r^s-10\cdot \eta \cdot \log_{10} (d/d_{near}) + \mathcal{N}(0,\sigma^2)$, 
where $r^s$ is a random sample for configuration $\mathcal{C}_{near}=(d_{near},$ $\theta_{near})$ such that $d_{near}=\argmin_{d_i \in \mathcal{D}} | d_i - d| $ and $\theta_{near}=\argmin_{\theta_i \in {\Theta}}$ $ | \theta_i - \theta_{rel} | $.
Note that we add an extra noise of variance $\sigma^2=2$ on top of the noisy samples (with $\sigma^2 \approx 4$) for configuration $\mathcal{C}_{near}$. 
To estimate the $\eta$,  we use \eqref{eqn:eta_est} to calculate $\eta_{ij}$ for each pair of distances, $d_i,d_j\in\mathcal{D}$ and take the average of them. 
We choose a value of $\delta t = 1s$ in \eqref{sec:state} to match the maximum achievable speed of our stepper motor as, ideally, the interval between any two consecutive movement control decisions could be $1s$ where the TrackBot carries out any movement control decision within the respective $1s$ interval.
}

\subsection{The Optimistic Strategy vs. The Pragmatic Strategy}
\label{sec:comparison_emulation}
\begin{figure}[!ht] 
 \centering
 \subfloat[Varying Leader Speed]{\label{fig:compare_maxvel_target}\includegraphics[width=0.8\linewidth,height=0.4\linewidth]{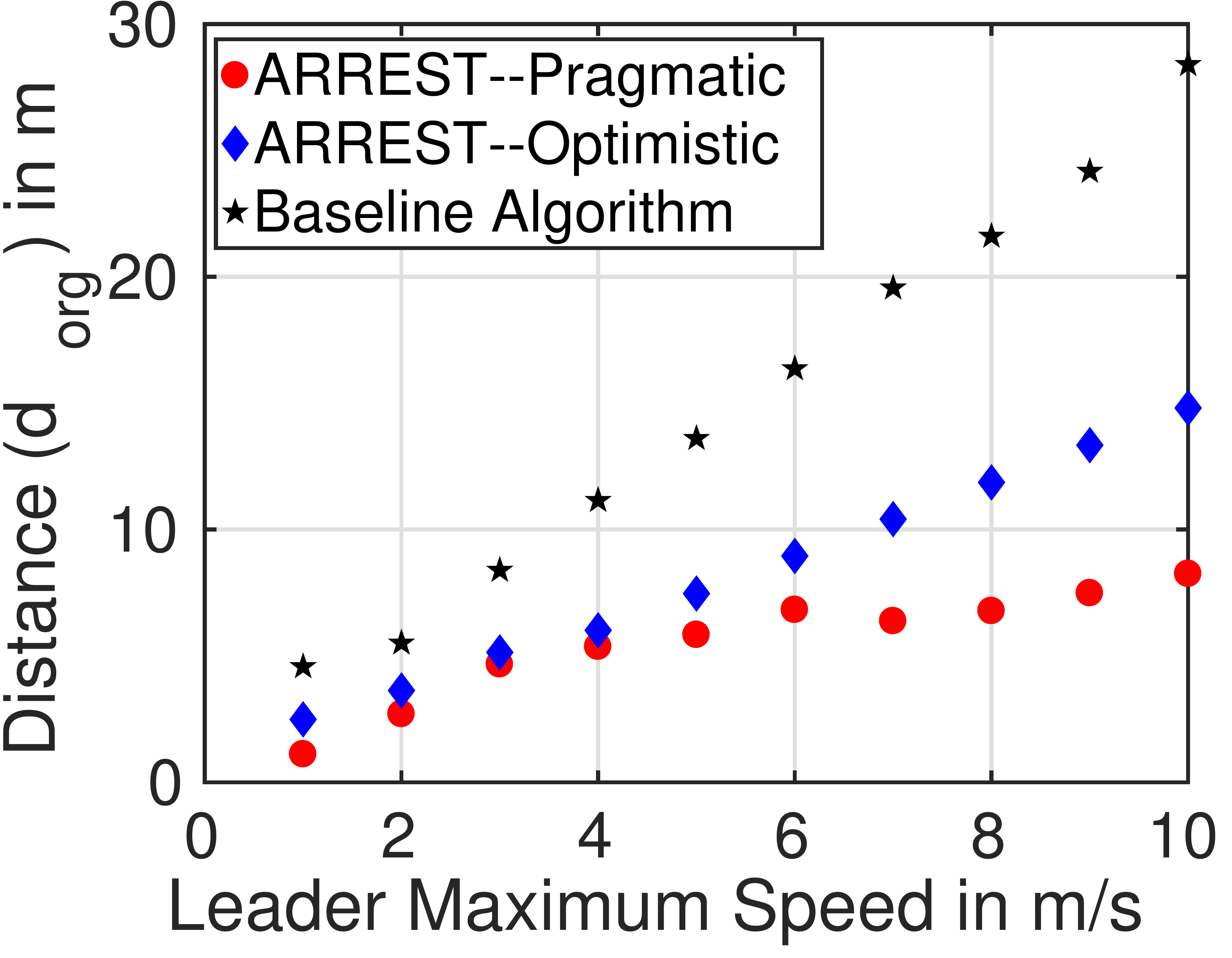}}\,
 \subfloat[Varying TrackBot's Speed]{\label{fig:compare_maxvel_tracker}\includegraphics[width=0.8\linewidth,height=0.4\linewidth]{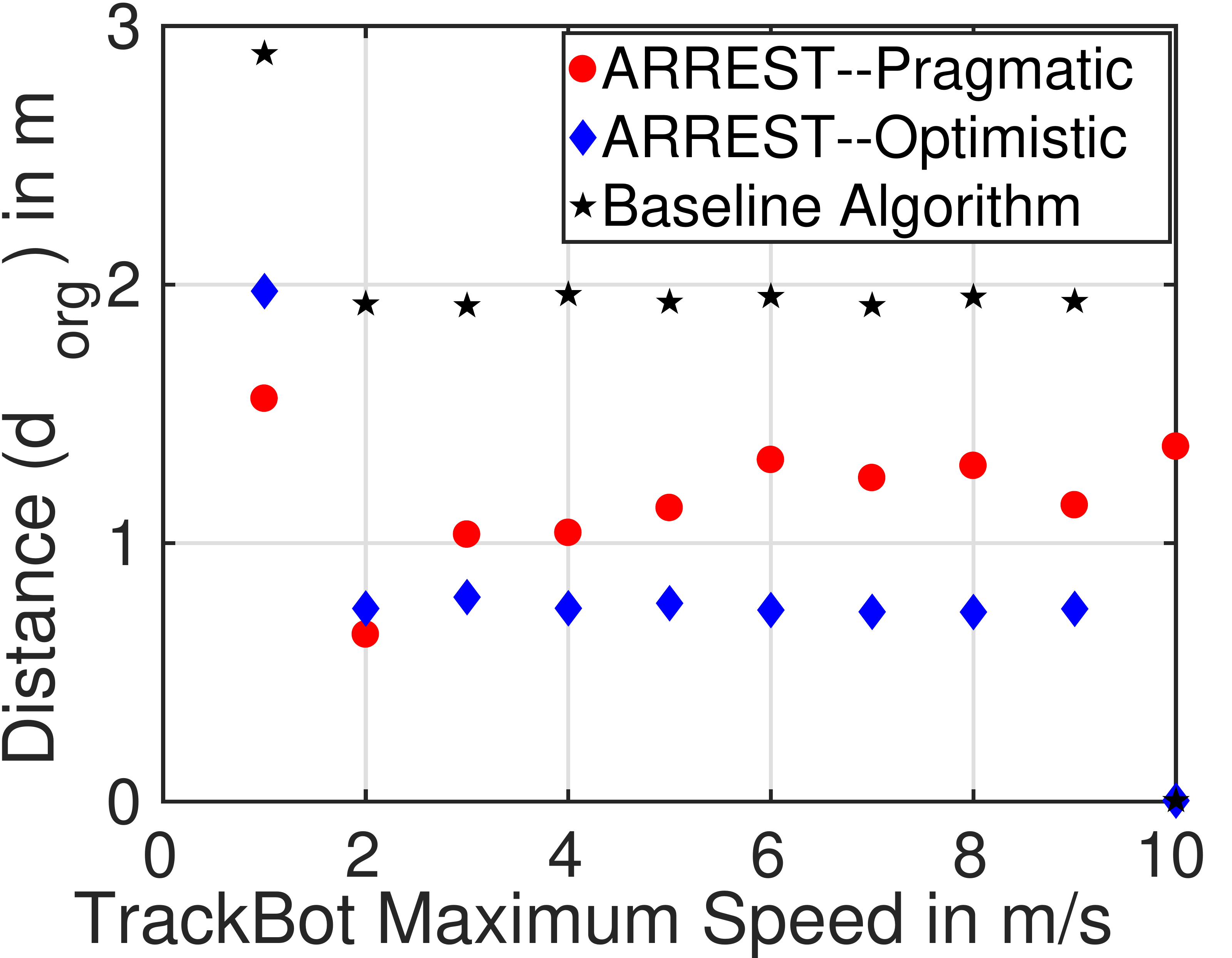}}
 \caption{  (a)-(b)Tracking Performance Comparison Among Different Speed Estimation Strategies}
 \label{fig:diff_path_dist}
\end{figure}

In this section, we compare the performance among the two proposed strategies, Optimistic and Pragmatic, and a Baseline algorithm. In the \textbf{Baseline algorithm}, the TrackBot estimates the relative position via the basic correlation method (discussed in~\ref{sec:aprox1}). Once the direction is determined, the TrackBot rotates to align itself toward the estimated direction and then moves with a speed of $\min\{v_{F}^{max}, \frac{d^e[n]}{\delta t}\}$. In Fig.~\ref{fig:compare_maxvel_target}, we compare the average distance between the TrackBot and the Leader for varying $v_{L}^{max}$ while setting $v_F^{max}=1.8\cdot v_{L}^{max}$. Figure~\ref{fig:compare_maxvel_target} clearly demonstrates that the Pragmatic strategy performs better than the Optimistic strategy as well as the Baseline algorithm, due to adaptability and accuracy of the speed information. The poor performance of the Optimistic strategy is due to its indifference towards the actual speed of the Leader which causes the TrackBot to lag behind for higher velocities. Conversely, we compare the average distance between the Leader and the TrackBot for varying $v_{F}^{max}$, while the Leader's maximum speed is fixed at $v_{L}^{max}=1m/s$. The experiment outcomes, presented in Fig.~\ref{fig:compare_maxvel_tracker}, show that the performance of both strategies are comparable, while the Optimistic strategy outperforms the Pragmatic strategy for $v_{F}^{max}\geq 3 \cdot v_{L}^{max}$. The reason behind this is the Leader is constantly changing movement direction while the TrackBot always travels along the straight line joining the last estimated position of the Leader and the TrackBot which may not be the same as the Leader's direction of movement. This results in oscillations in the movement pattern for the Pragmatic strategy while the Optimistic strategy avoids oscillations since it assumes the Leader to be static. The worst performance of the Baseline approach is attributed to lack of speed adaptation by taking past observation into account.  

One more noticeable fact from Fig.~\ref{fig:compare_maxvel_tracker} is that if $v_{F}^{max}=v_{L}^{max}$, the tracking performance is the worst. This is quite intuitive because for this speed configuration, the TrackBot is unable to compensate for any error or initial distance while the Leader constantly moves at a speed close to $v_{L}^{max}$. Thus, the relative speed needs to be positive for proper tracking. In order to find a lower bound on the TrackBot's speed requirement, we perform another set of experiments by varying $v_F^{max}$ from $v_{L}^{max}$ to $3\cdot v_{L}^{max}$. {Based on the results, we conclude that for $v_{F}^{max} \leq 1.6\cdot v_L^{max}$, the tracking system fails and the distance increases rapidly.} On the other hand, for $v_{F}^{max}>1.6\cdot v_{L}^{max}$ the performance remains the same. Thus, in our experimental setup, we opt for $v_{F}^{max}=1.8\cdot v_{L}^{max}$. 


\subsection{Absolute Distance Statistics}
One main focus of our ARREST architecture is to guarantee {\small$\mathbb{P}\left(||\mathbf{X}_L[n]\right.$ $\left.-\mathbf{X}_F[n]||_2 \leq D_{th}\right) \approx 1 \ \  \forall n$}.
The value of $D_{th}$ could be chosen as a function of $v_{L}^{max}$. However, according to our target application context, we select $D_{th}= 5m$ as we consider a distance more than 5 meters to be large enough to lose track in an indoor environment. With this constraint, we find that our present implementation of the ARREST system fails in the tracking/following objective if the Leader moves faster than 3m/s.
In order to verify whether our ARREST architecture can guarantee the distance requirement for Leader with $v_{L}^{max}\leq 3m/s$,  we perform a set of emulations with $\delta t=1s$, where the Leader travels along a set of random paths. 
In all cases, the instantaneous distances between the TrackBot and the Leader during the emulation are less than $5m$ with probability $\approx 1$. The nonzero probability of distances higher than $5m$ is due to randomness in the Leader's motion including complete reversal of movement direction.

\begin{figure}[!ht] 
 \centering
 \subfloat[]{\label{fig:emu_error_dist}\includegraphics[width=0.36\linewidth,height=0.4\linewidth]{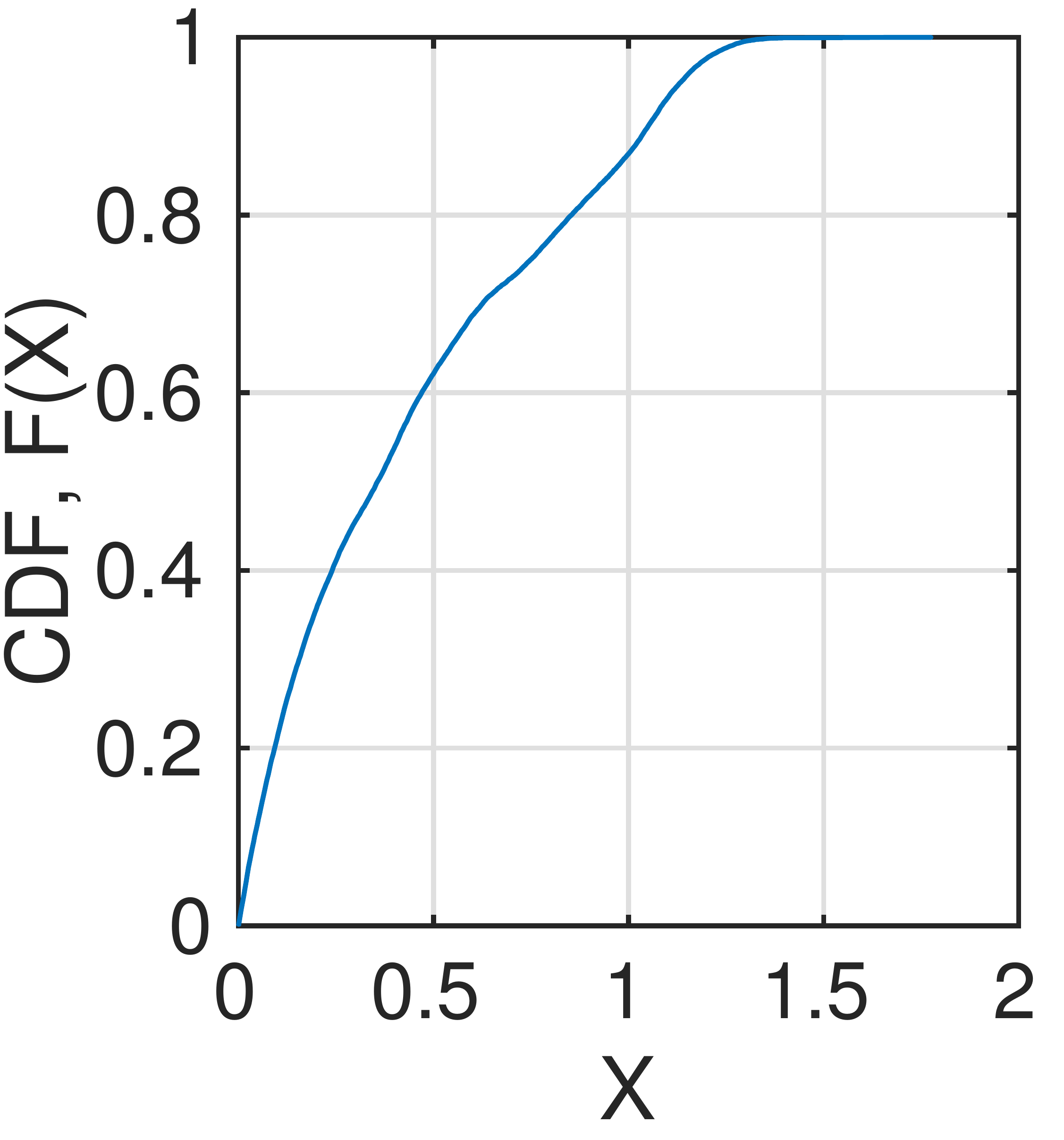}}
 \subfloat[]{\label{fig:emu_error_angle}\includegraphics[width=0.32\linewidth,height=0.4\linewidth]{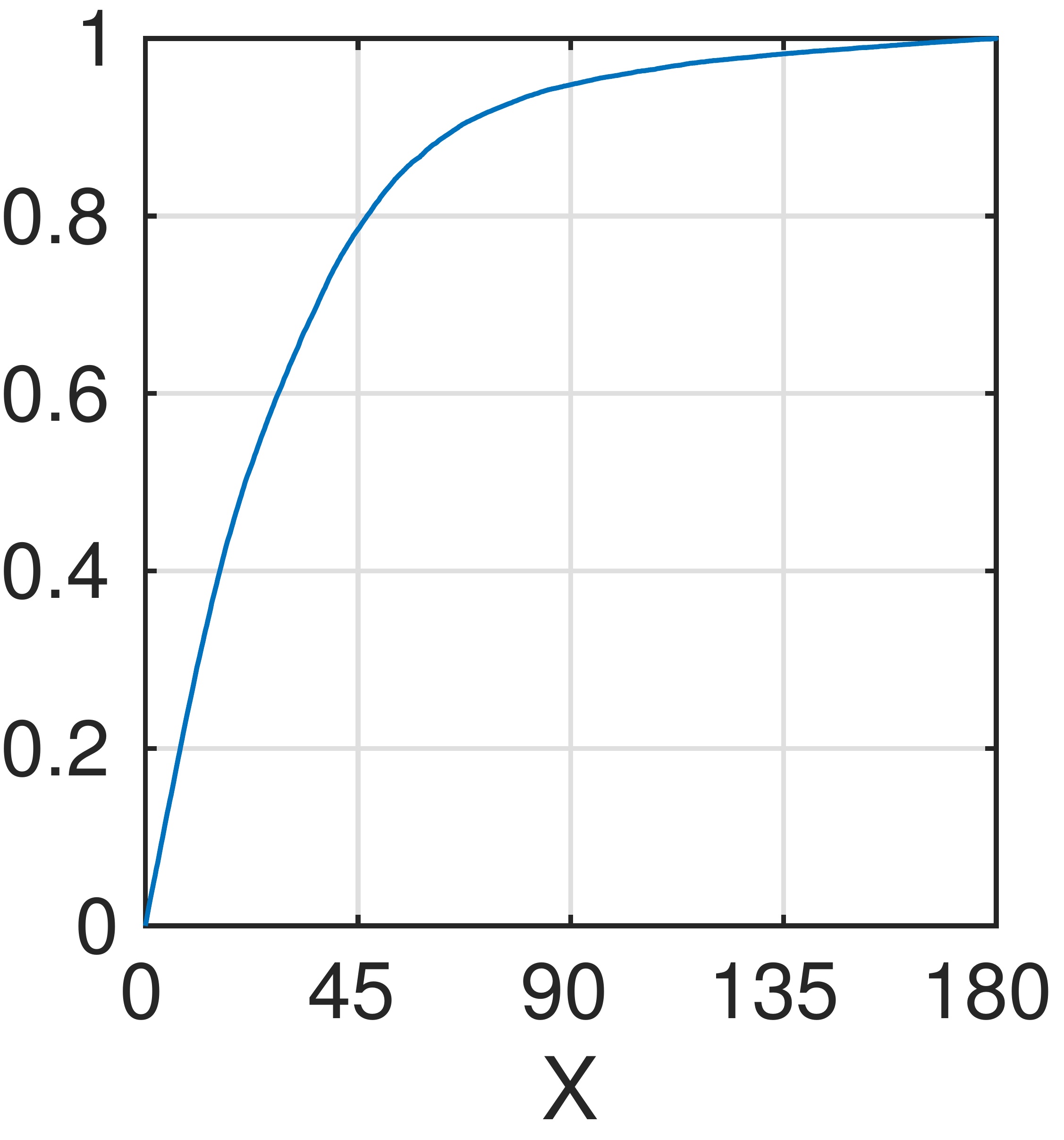}}
 \subfloat[]{\label{fig:emu_error_speed}\includegraphics[width=0.32\linewidth,height=0.4\linewidth]{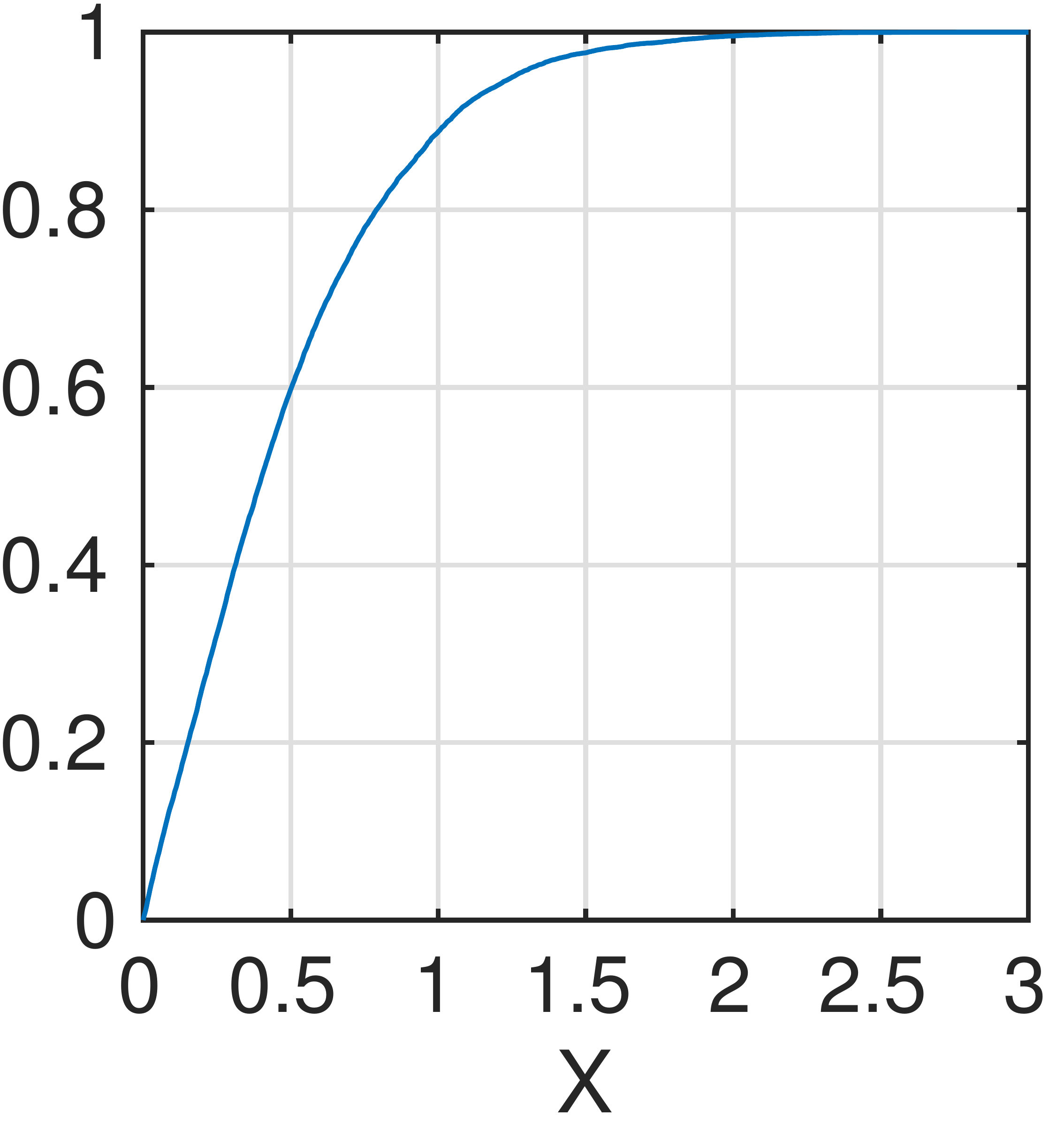}}
 \caption{Emulation Based Performance: (a) Absolute Distance Estimation Errors (in m), (b) Absolute Angle Estimation Errors (in degrees), and (c) Absolute Speed Estimation Errors (in m/s) }
 \label{fig:error_dist1}
\end{figure}
\begin{figure*}[!ht] 
 \centering
\subfloat[]{\label{fig:diste_stat_in_real}\includegraphics[width=0.22\linewidth,height=0.22\linewidth]{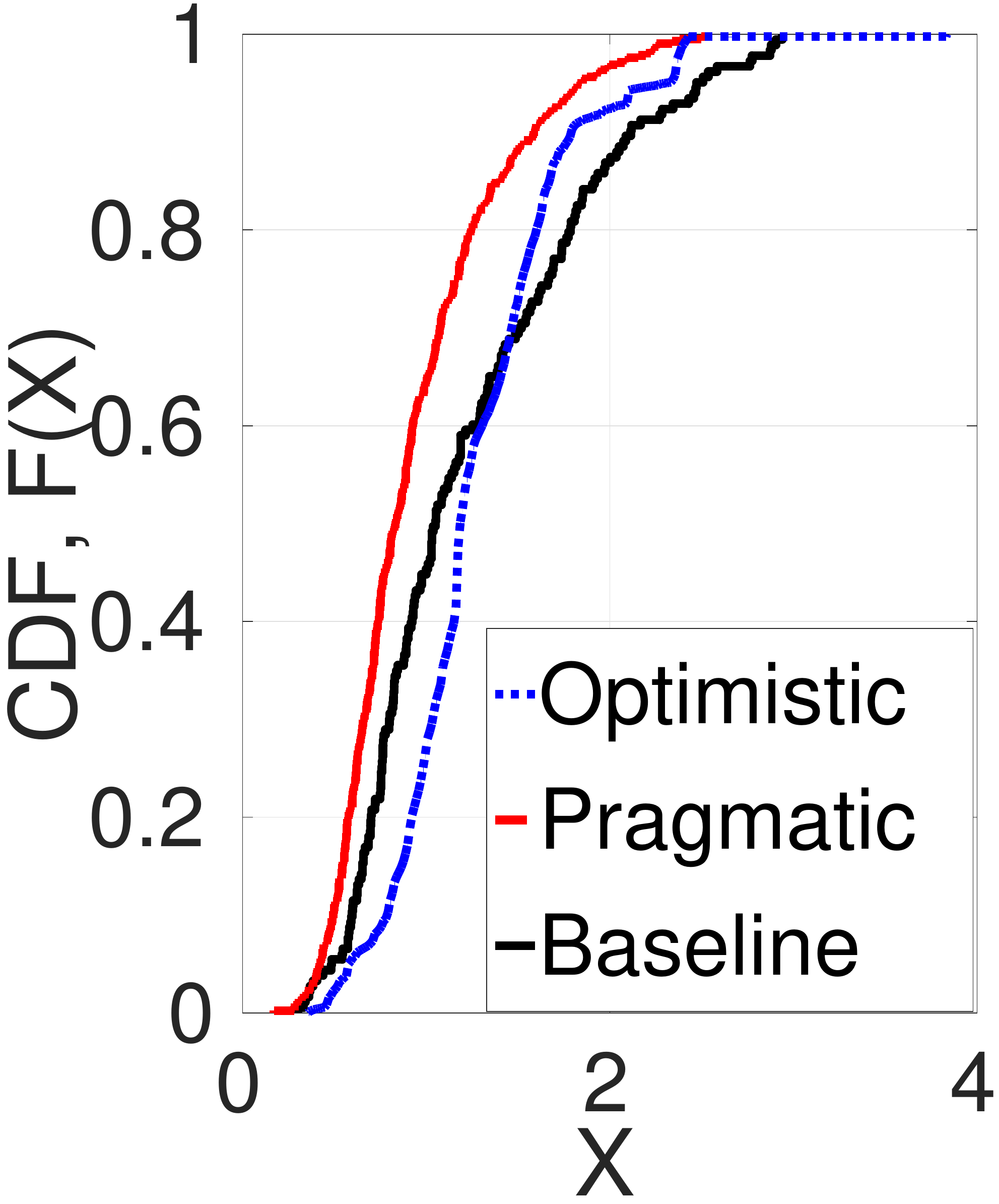}}\qquad
\subfloat[]{\label{fig:dist_error_stat_in_real}\includegraphics[width=0.22\linewidth,height=0.22\linewidth]{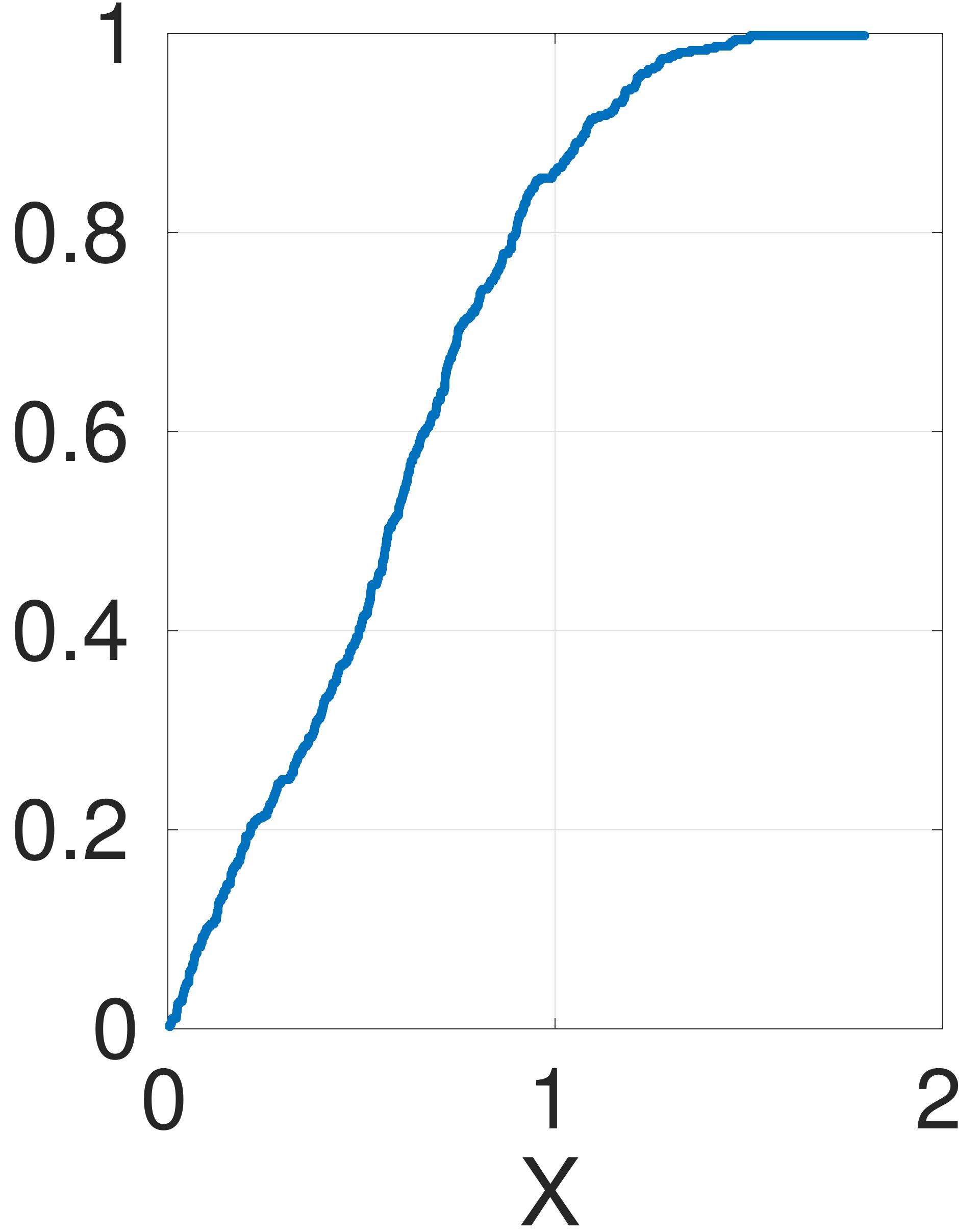}}\qquad
\subfloat[]{\label{fig:angle_error_stat_in_real}\includegraphics[width=0.22\linewidth,height=0.22\linewidth]{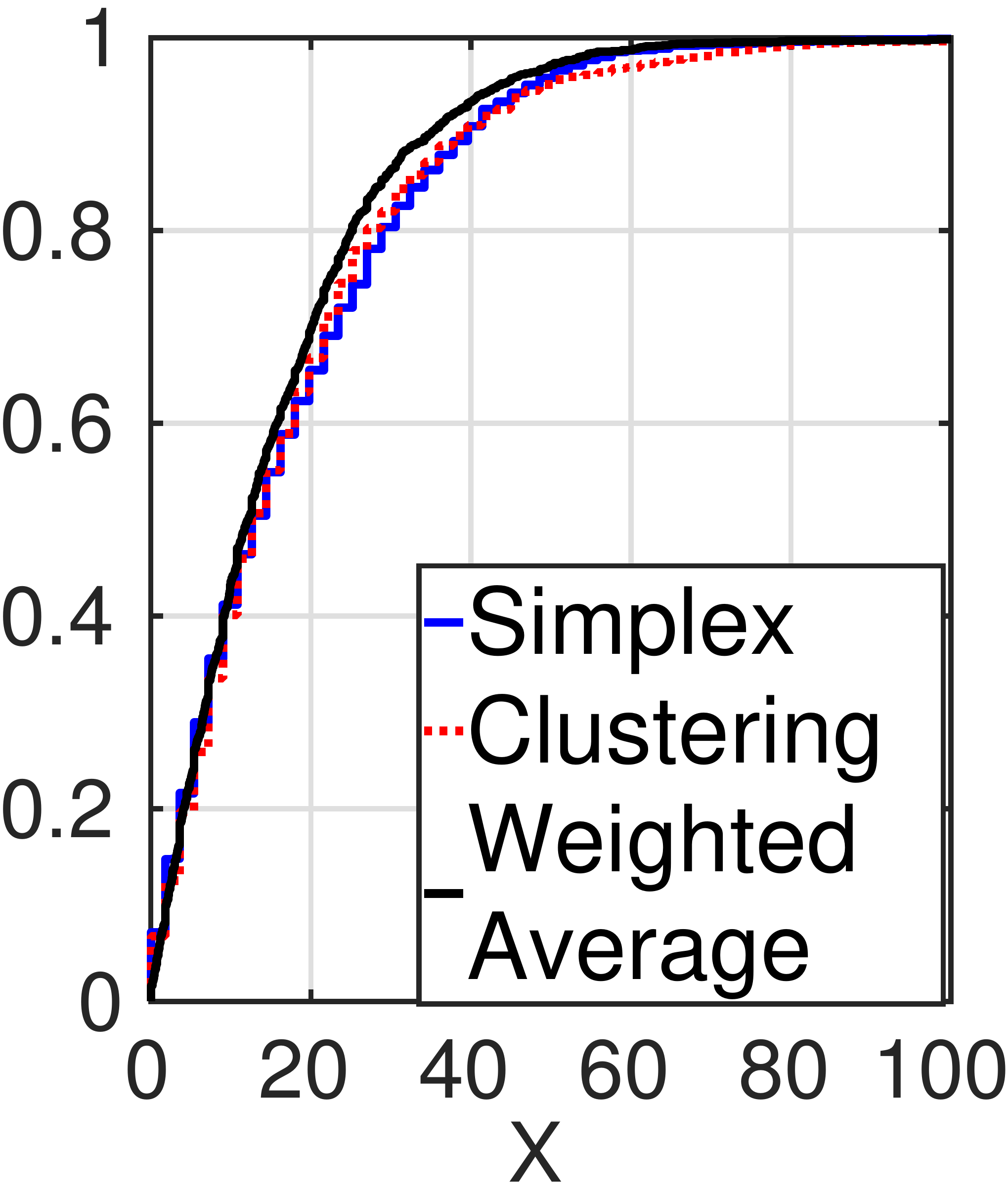}}
 \caption{ Real Experiment Based Performance for Small Scale: (a) Absolute Distance in Meters, (b) Absolute Distance Estimation Error in Meters, and (c) Absolute Angle Estimation Error in Degrees}
 \label{fig:error_dist_practical}
\end{figure*}
\subsection{Estimation Errors} In order to learn the statistics of different
estimation errors, we perform a range of emulation experiments, where the Leader
follows a set of random paths and $v_{L}^{max}\leq 3m/s$.  In
Fig.~\ref{fig:emu_error_dist}, we plot the empirical CDF of the absolute errors
in the distance estimates maintained by our system.  Figure~\ref{fig:emu_error_dist}
clearly illustrates that the instantaneous errors are less than $100cm$ with
very high probability ($\approx 90\%$), and that the absolute error values are
bounded by $1.5m$. These statistics are reasonable for pure RSSI-based
estimation systems (explained further in Section~\ref{sec: raw}).  We also plot
the CDF of the absolute angle estimation errors over the duration of the
emulations in Fig.~\ref{fig:emu_error_angle}. It can be seen that the
absolute angle errors are less than $40^\circ$ with high ($\approx 80\%$)
probability, which is justified as the Half Power Beam Width (HPBW) for the
antenna we are using is approx $70^\circ$. Further improvements may be possible
by using an antenna with greater directionality or other radios (such as UWB
radios). The non-zero probability of the angle error being more than $40^\circ$
is again due to the random direction changes in the Leader's movements.
Similarly, we analyze the absolute speed estimation errors in terms of CDF,
illustrated in Fig.~\ref{fig:emu_error_speed}. The absolute errors in the
speed estimations of the Leader are less that $1m/s$ with $\approx 90\%$
probability.

{\small
\begin{table}[!h]
    \centering
        \caption{Summary of Emulation Results}
    \resizebox{\linewidth}{!}
    {\small
    \begin{tabular}{|m{\linewidth}|}
    \hline
        $\Box$ Pragmatic Strategy performs best for $1.6 \cdot v_{L}^{max} < v_{F}^{max} <3\cdot  v_{L}^{max}$ while Optimistic Strategy performs best for $v_{F}^{max}\geq 3\cdot v_{L}^{max}$\\
        $\Box$ The ARREST system fails if  $v_{L}^{max}>3m/s$.\\
        $\Box$ For $v_{L}^{max}\leq 3m/s$ and $v_{F}^{max}=1.8 \cdot v_{L}^{max}$, the TrackBot stays within $5m$ of the Leader with probability $\approx100\%$.\\
        $\Box$ Absolute distance estimation errors are $<100cm$ with probability $\approx90\%$ and $<150cm$ with probability $\approx 100\%$.\\ 
        $\Box$ Absolute angle estimation errors are $<40^\circ$ with  probability $\approx 80\%$.\\
        $\Box$ Absolute speed estimation errors are less than $1m/s$ with probability $\approx 90\%$. \\
        \hline
    \end{tabular}
    }
    \end{table}
}

\section{Real Experiment Results : Small Scale}
\label{sec:real_exp}
To analyse the performance of the ARREST architecture, we use the TrackBot prototype to perform a set of small scale experiments, followed by a range of large scale experiments. In this section, we present the results of our small-scale real-world experiments. 

\subsection{Method} 
\begin{figure*}[!ht] 
 \centering
  \subfloat[Indoor]{\label{fig:indoor_trace}\includegraphics[width=0.32\linewidth, height=0.2\linewidth]{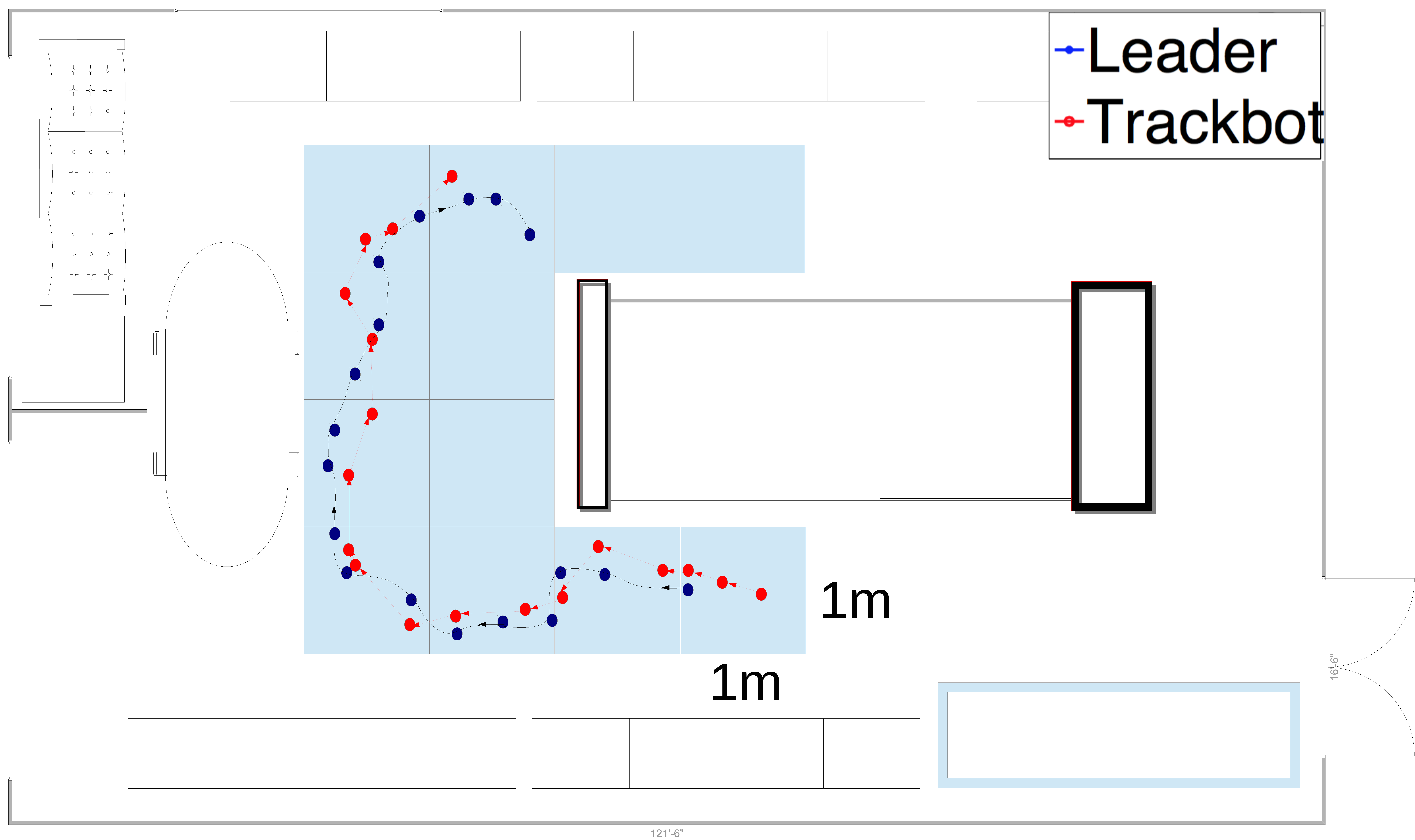}} \qquad
  \subfloat[Hallway]{\label{fig:hallway_trace}\includegraphics[width=0.32\linewidth, height=0.22\linewidth]{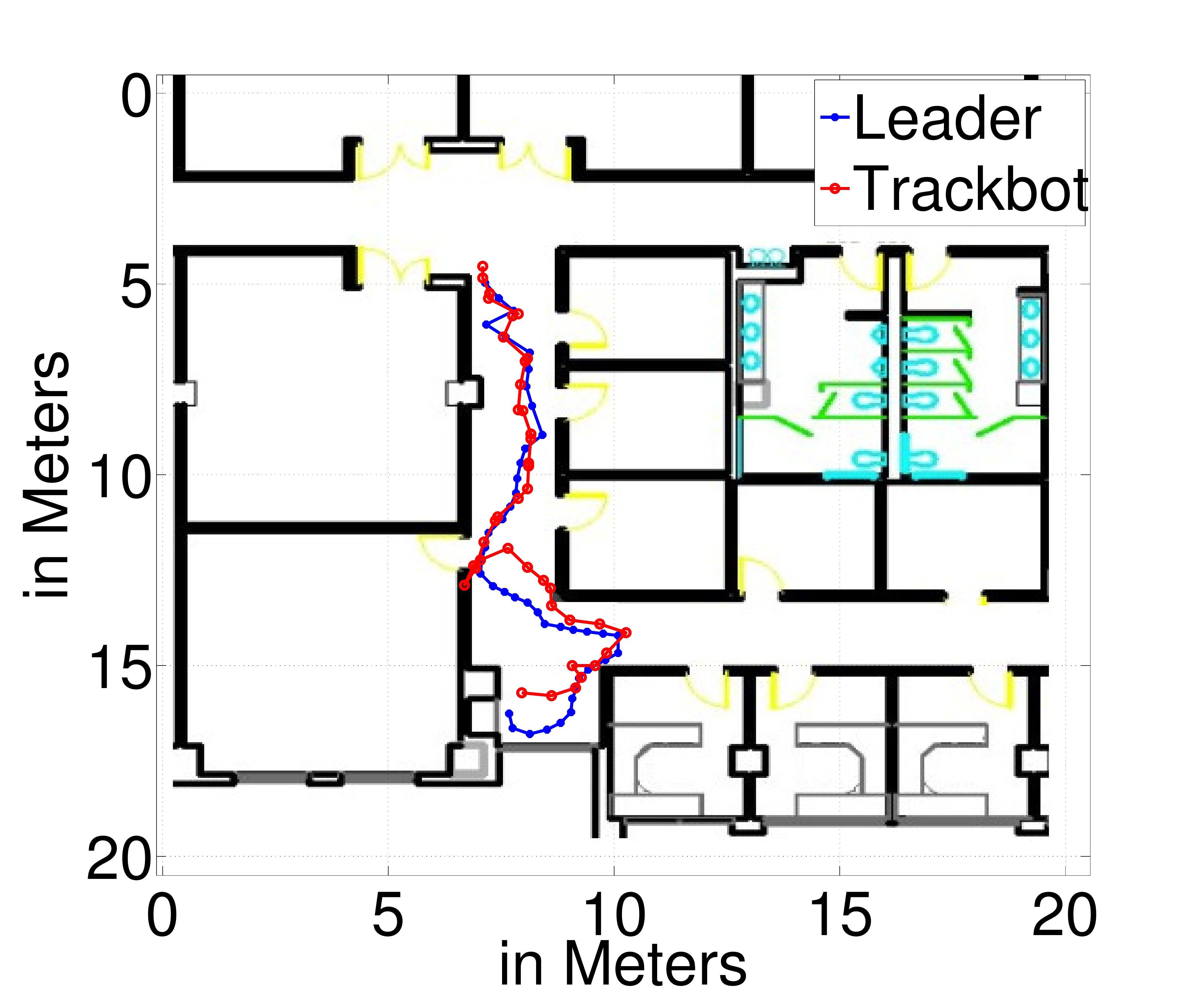}}\, 
  \subfloat[Indoor (No Line of Sight)]{\label{fig:indoor_static}\includegraphics[width=0.32\linewidth, height=0.2\linewidth]{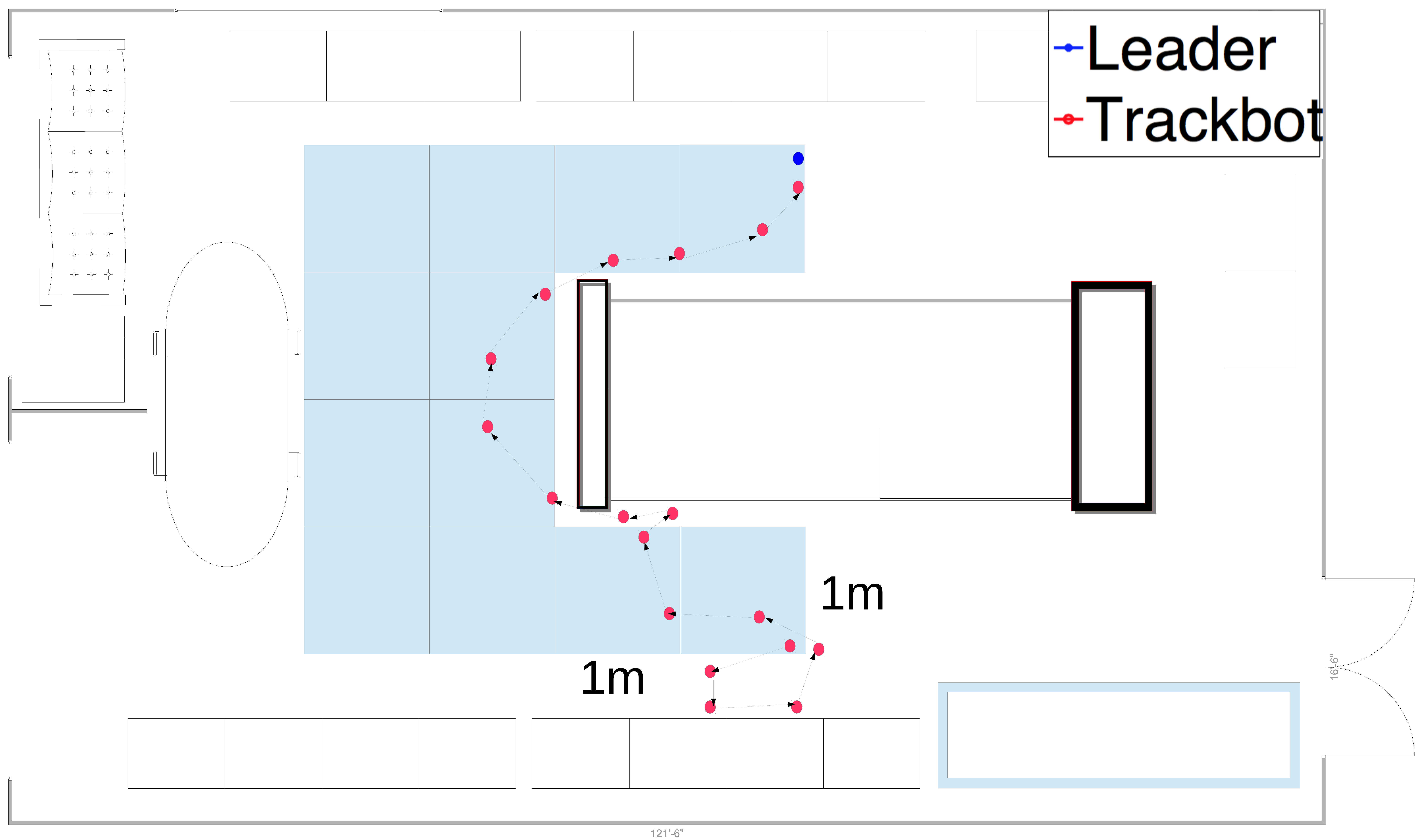}} \qquad
  \subfloat[VICON System]{\label{fig:vicon_trace}\includegraphics[width=0.32\linewidth, height=0.22\linewidth]{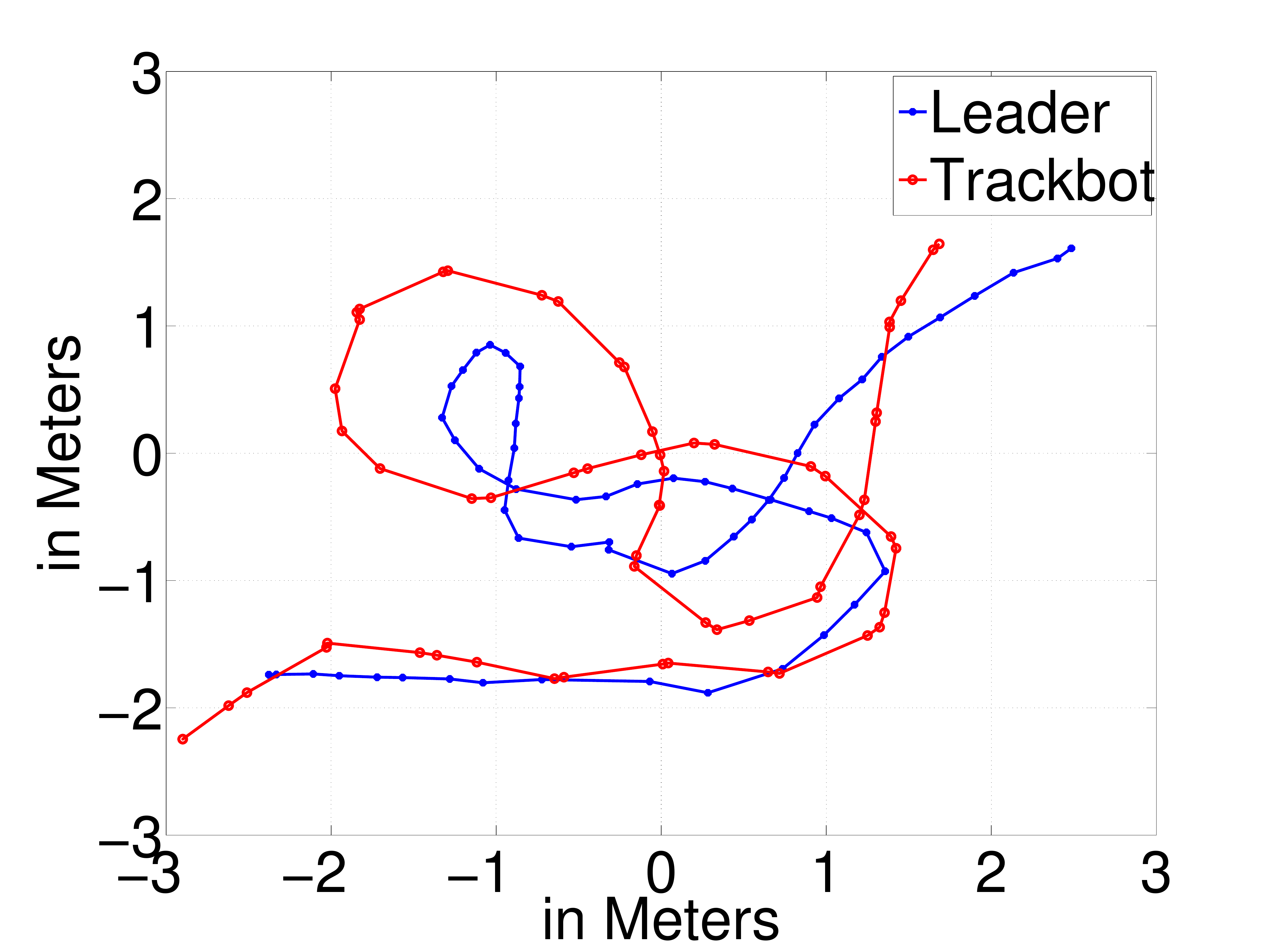}}
 \caption{Full Path Traces from Small Scale Real World Experiments}
 \label{fig:practical_trace}
\end{figure*}

Based on the valuable insights from the emulation results, we choose TrackBot's speed to be at least 1.8X the Leader's speed. 
The TrackBot makes a decision every $6s$. Between each decision, the TrackBot takes $2s$ for both the antenna rotation and RSSI scan, $2s$ for the chassis rotation, and $2s$ for the chassis translation. However, in the state update equations, $\delta t=4s$ because the actual chassis movement takes place for only $4s$.
With this setup, we perform a set of real tracking experiments in three different environments:

$\Box $ A cluttered office space, illustrated in Fig.~\ref{fig:indoor_trace} ($\approx 10m \times 6m$), with a lot of office desks, chairs, cabinets, and reflecting surfaces.

$\Box $ A hallway, illustrated in Fig.~\ref{fig:hallway_trace} ($\approx 18m$ long and $3m$ wide), with pillars as well as sharp corners.

$\Box $ A VICON camera localization~\cite{vicon} based robot experiment facility, illustrated in Fig.~\ref{fig:vicon_trace} ($\approx 6m\times 6m $). 

For the first two environments, we use manual markings on the floor to localize both the Leader and the TrackBot. For the last environment, the VICON facility provides us with camera-based localization at millimeter scale accuracy. 
We perform a set of experiments in each of these environments for an approximate total period of one month with individual run lasting for $30$ minutes during different times of the day. For these experiments, the Leader is a human carrying an OpenMote transmitter.


\subsection{The Optimistic Strategy vs. The Pragmatic Strategy}
\label{sec:comparison_real_world}
Similar to our emulation based analysis, we perform a real system based comparison of the
proposed speed adaptation strategies as well as the \textbf{Baseline Algorithm} (introduced in Section~\ref{sec:comparison_emulation}).
However, in this set of experiments we do not vary the maximum speed of the TrackBot or the Leader due to prototype hardware limitations.
Instead, we compare the absolute distance CDF statistics of these three strategies in Fig.~\ref{fig:diste_stat_in_real} for $v_{F}^{max}=10cm/s$ and $v_{F}^{max}=1.8\cdot v_{L}^{max}$. 
Figure~\ref{fig:diste_stat_in_real} validates that Pragmatic strategy performs best among all three strategies when $v_{F}^{max}=1.8\cdot v_{L}^{max}$. Moreover, the baseline strategy performs the worst due to lack of speed adaptation as well as lack of history incorporation. In summary, our real experiment based results concur with the emulation results.

\subsection{Estimation Errors}
\label{sec:real_est_err}
To analyze the state estimation errors in our ARREST architecture similar to the emulations, we perform a range of prototype based experiments, where the $v_{F}^{max} = 1.8\cdot v_{L}^{max}$ and the Leader follows a set of random paths. 
In Fig.~\ref{fig:dist_error_stat_in_real}, we plot the empirical CDF of the absolute errors in the distance estimates maintained by our TrackBot. 
Figure~\ref{fig:dist_error_stat_in_real} clearly illustrates that the instantaneous absolute errors in our distance estimates are $\leq 100cm$
with very high probability ($\approx90\%$), and are bounded by $1.5m$. These statistics are also reasonable for pure RSSI based estimation systems and concur with the emulation results.
Next, in Fig.~\ref{fig:angle_error_stat_in_real}, we compare the angle estimation error performance of the TrackBot for all three AoA observation methods introduced in Section~\ref{sec:arrest_angle} where we intentionally introduce random sparsity in the RSSI measurements.  
Figure~\ref{fig:angle_error_stat_in_real} illustrates that our proposed \textbf{clustering method} and \textbf{weighted average method}
perform significantly better than the \textbf{basic correlation method} which is expected since the first two take into account the clustered sparsity (Detailed in Section~\ref{sec:arrest_angle}).
The instantaneous absolute angle errors are less than $40^\circ$ with high probability ($\approx 90\%$) for all three methods which is justified because the HPBW specification for the antenna is approx $70^\circ$.
Figure~\ref{fig:angle_error_stat_in_real} also illustrates that the weighted angle observation method slightly outperforms the clustering method for AoA observation.
The apparent similarity between the performance of the {clustering method} and the {weighted average method} is attributed to the consistent lower cluster sizes compared to the gap sizes ($\lambda_{a} << \mu_{a}$) in our experiments. 


\subsection{Tracking Performance}
In Fig.~\ref{fig:indoor_trace}, we present a representative path trace from the experiments in the indoor scenario. 
Similarly, in Fig.~\ref{fig:hallway_trace} we present a real experiment instance in the Hallway. 
Lastly, Fig.~\ref{fig:vicon_trace} illustrates an example trace from the VICON system. 
All three figures illustrate that our system performs quite well in the respective scenarios and stays within $\approx 2m$ from the Leader for the duration of the experiments. 
These results suggest that our system works equally well in different environments: cluttered and uncluttered. 
To verify that further, we perform a set of experiments with a static Leader not in the line of sight of the TrackBot for $\geq 50\%$ of the TrackBot's path. 
\emph{Our TrackBot was able to find the Leader in $75\%$ of such experiments. }
In Fig.~\ref{fig:indoor_static}, we present one instance of such experiment. 
The main reason behind this success lies in the TrackBot's ability to leverage a good multipath signal (if exists). 
In absence of direct line of sight, the TrackBot first follows the most promising multipath component and by doing so it eventually comes in line of sight with the Leader and follows the direct path from that point on.
\emph{In most of these experiments ($\geq 90\%$), the TrackBot travels a total distance of less than 2X the distance traveled by the Leader. 
This implies that our system is efficient in terms of energy consumption due to robotic maneuvers. }

Nonetheless, these small real-world experiments also point out that our current system does not work if there exists no strong/good multipath signal in NLOS situations where ``strong multipath'' implies that one multipath signal's power is significantly higher than other multipath signals. We detail multipath related problems and our method of partly circumventing it in Section~\ref{sec:multipath_adapt}.

{\small
\begin{table}[!h]
    \centering
    \caption{Summary of Small Scale Real-World Experiments}
    \resizebox{\linewidth}{!}
    {\small
    \begin{tabular}{|m{\linewidth}|}
    \hline
    $\Box$ Pragmatic Strategy performs best for $1.8 \cdot v_{L}^{max} = v_{F}^{max}$.\\
    $\Box$ Absolute distance estimation errors are $<100cm$ with probability $\approx90\%$ and $<150cm$ with probability $\approx 100\%$.\\ 
    $\Box$ Absolute angle estimation errors are $<40^\circ$ with  probability $\approx 90\%$.\\
    $\Box$ Weighted average AoA observation method performs the best.\\
    $\Box$ The TrackBot stays within $2m$ of the Leader with probability $\approx98\%$ in line of sight contexts.\\
    $\Box$ The ARREST system works with probability $\approx 75\%$ for NLOS contexts, although it fails if no ``strong multipath'' exists.\\
    \hline
    \end{tabular}
    }
    \end{table}
}

\section{Real Experiment Results : Large Scale}
\label{sec:real_exp_large}

\subsection{Method} 
The small scale experiments, presented in
Section~\ref{sec:real_exp}, were limited in terms of deployment region ($\leq
60$ sq. meters) due to the dimensions of the VICON system and the effort plus
time required for large scale experiments with manual measuring/markings. To
perform large scale and long duration experiments based evaluations, we
integrated a version of a well known Time Difference of Arrival based
localization~\cite{savvides2001dynamic,maloney2000enchanced} ground truth system
in our TrackBot. This helped us avoid the need of tedious manual markings and
measurements. For more efficient experiments, we also developed a robotic
leader, which we will refer to as the \textbf{LeaderBot} in this section, to act
as both Leader as well as the reference node for the TDoA localization system.

The main idea behind TDoA systems is to use a reference node that transmits two different types of signals, say RF and Ultrasound, simultaneously. Now, the localizing/receiver node receives these two signals at different instances of time due the propagation speed difference between RF and Ultrasound, say $\Delta c$. With proper timestamps, the receiver can now calculate the time difference of arrival of these two signal, say $\Delta t$, to estimate the distance as $\Delta c \cdot \Delta t$. We extend this concept slightly further by placing both the receiver RF antenna and the ultrasound on the the TrackBot's rotating platform. We rotate the platform in steps of $18^\circ$ (just a design choice) and perform TDoA based distance estimation for each orientation of the assembly. The TDoA system returns a valid measurement if and only if the assembly is oriented toward a direct line of sight or a reflected signal path. Assuming that there exists a line of sight, the orientation with the smallest TDoA corresponds to the actual angle between the LeaderBot and the TrackBot, and value of the smallest TDoA corresponds to the distance. 

\subsection{LeaderBot and TDoA Ranging} 
\begin{figure*}[!ht] 
 \centering
\subfloat[]{ \label{fig:3pi}\includegraphics[width=0.15\linewidth, height=0.25\linewidth]{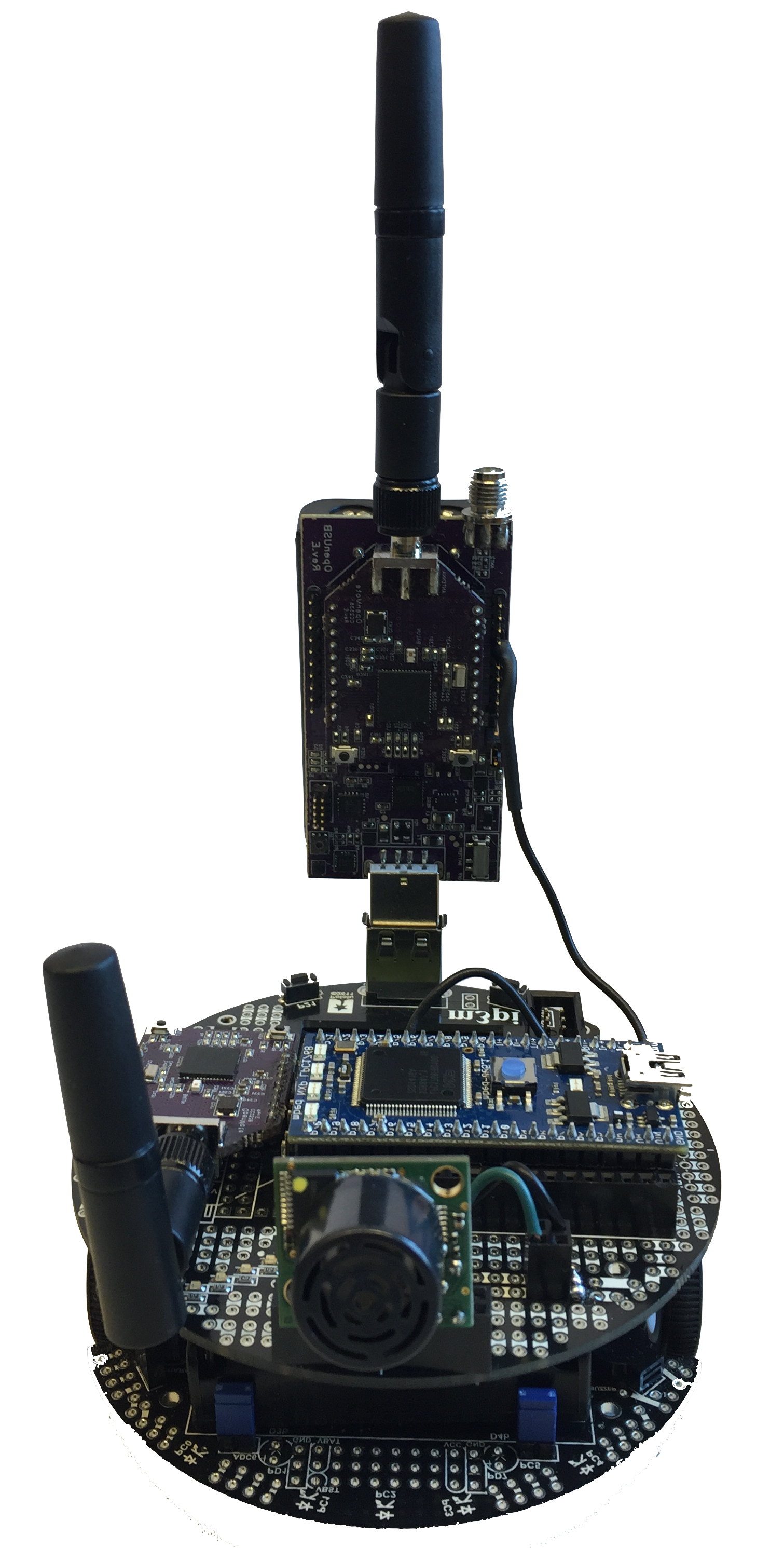}}
\subfloat[]{\label{fig:practical_large_env_dist}\includegraphics[width=0.25\linewidth,height=0.25\linewidth]{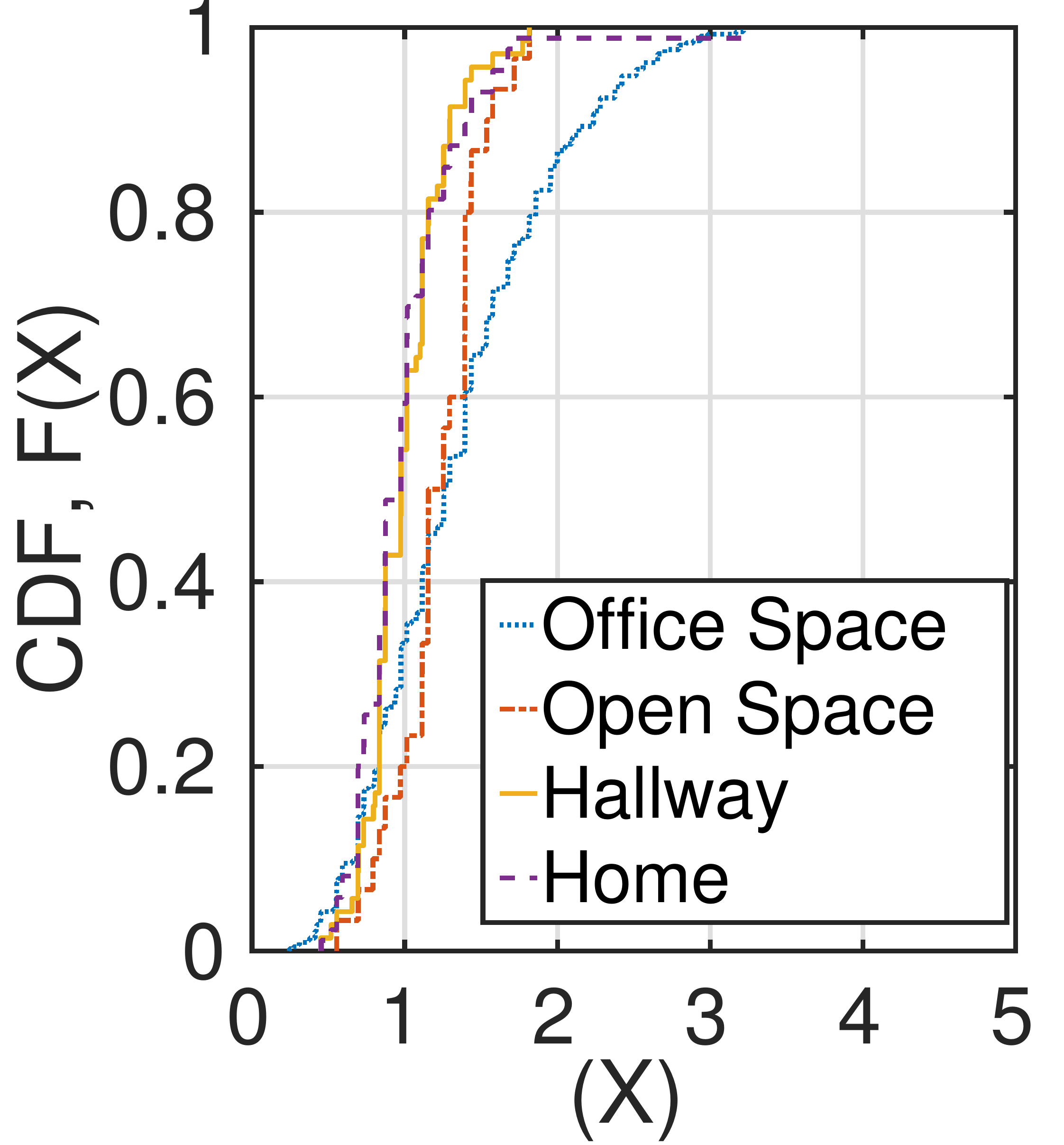}}\qquad
\subfloat[]{\label{fig:practical_large_env_dist_error}\includegraphics[width=0.23\linewidth,height=0.25\linewidth]{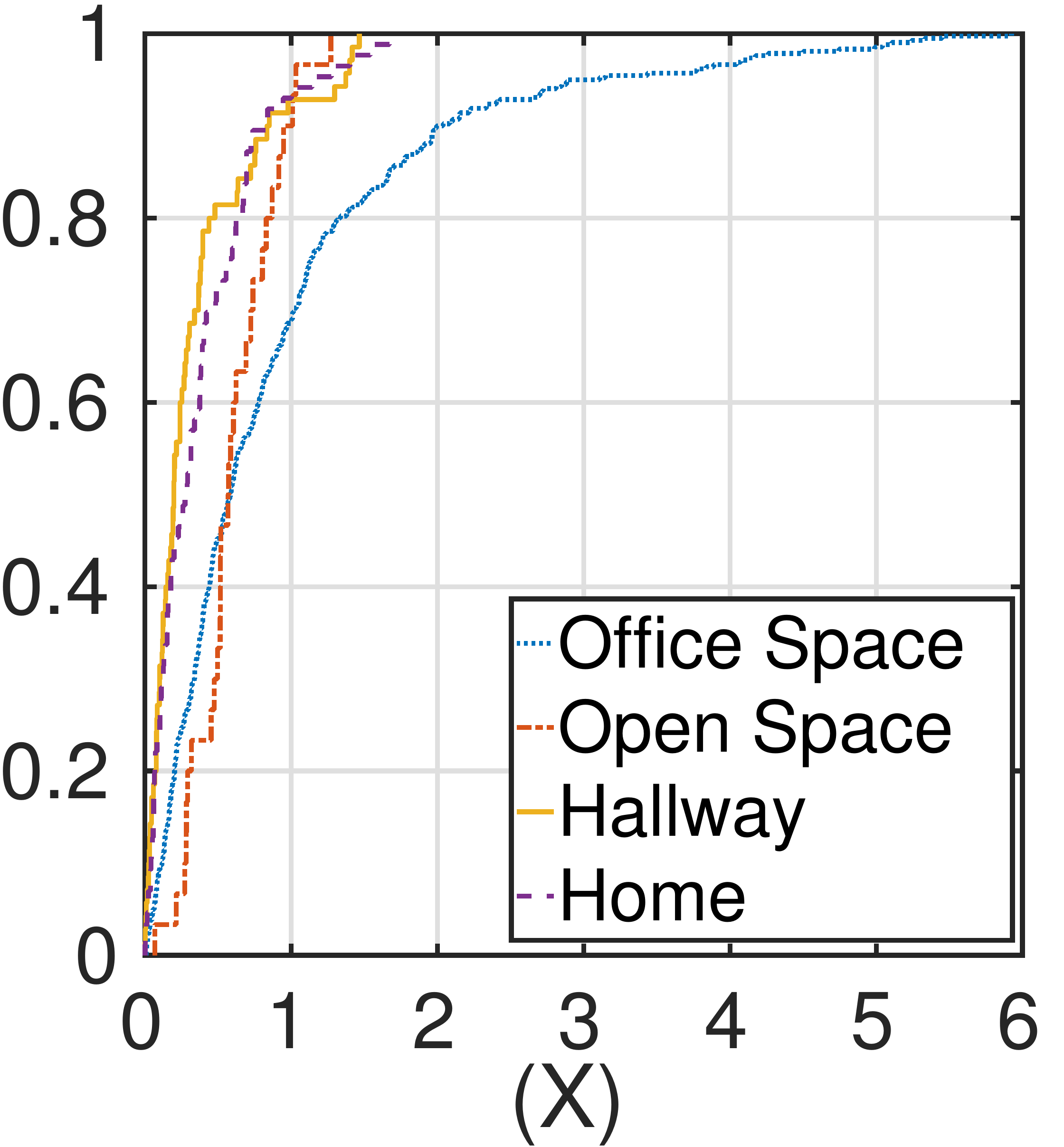}}\qquad
\subfloat[]{\label{fig:practical_large_env_angle_error}\includegraphics[width=0.23\linewidth,height=0.25\linewidth]{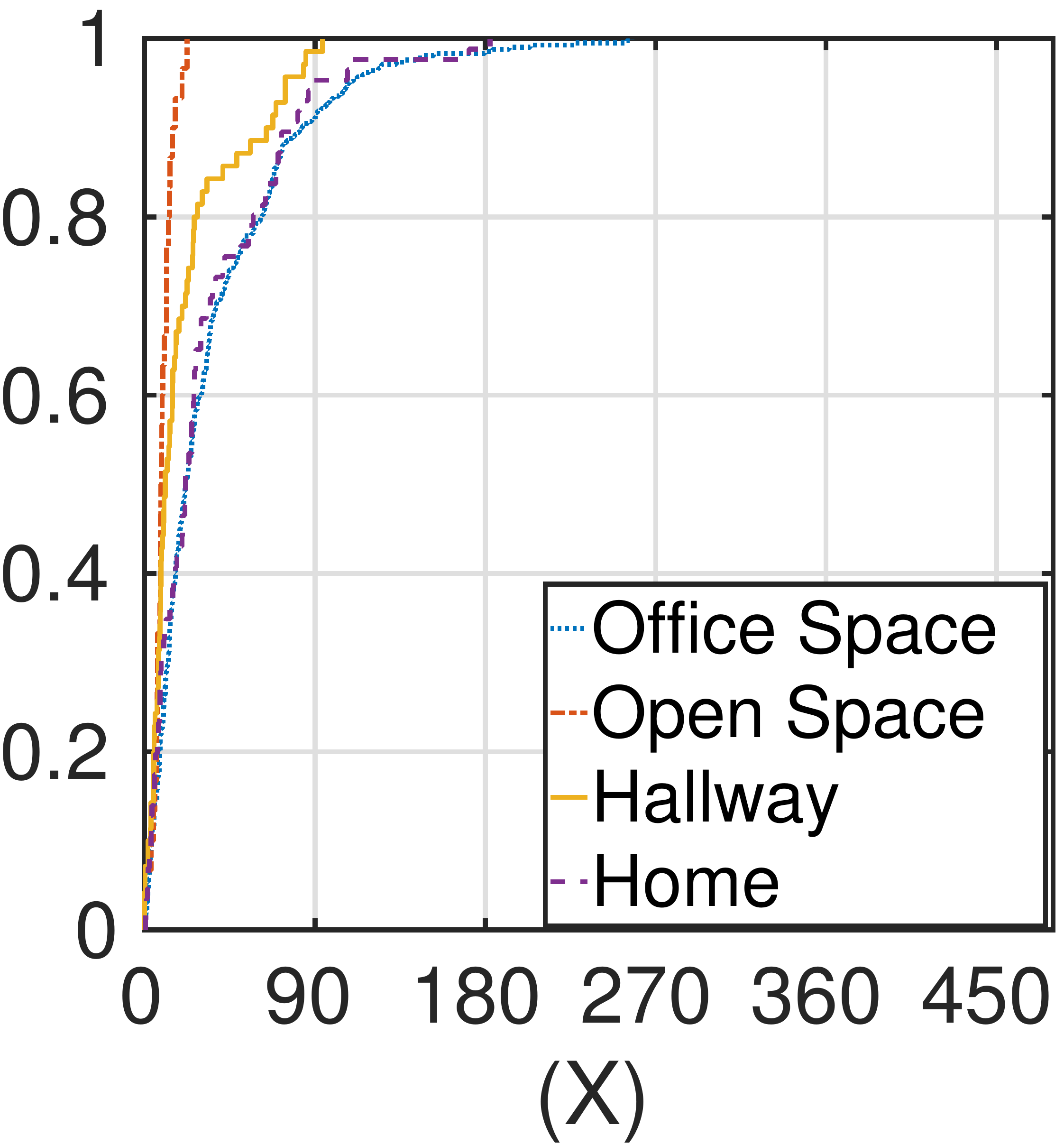}}

 \caption{ Real Experiment Based Performance for Large Scale: (a) 3pi LeaderBot (b) Absolute Distance in Meters, (c) Absolute Distance Estimation Error in Meters, and (d) Absolute Angle Estimation Error in Degrees}
 \label{fig:error_dist_practical_large_env  (b) }
\end{figure*}

The LeaderBot is built upon the commercially available small Pololu 3pi robot~\cite{pololu}. 
\emph{In our LeaderBot, we use two Openmotes: one Openmote acts as the Leader beaconer \textbf{(Beacon Mote)} and operates on 802.15.4 channel 26; the other Openmote \textbf{(Range Mote)} is used to remotely control the 3pi robot's movements and to perform the TDoA based localization on 802.15.4 channel 25.} We use two Openmotes for cleaner design as well as to avoid operation interference between remote controlling and beaconing. We use a MB 1300 XL-MaxSonar-AE0~\cite{mbultra} as the ultrasound beaconer, powered by the 3pi robot. The LeaderBot is illustrated in Fig.~\ref{fig:3pi}.
On the TrackBot, we also add a MB 1300 XL-MaxSonar-AE0~\cite{mbultra} ultrasound on the rotating platform along side with the directional antenna to receive the ultrasound beacons. 
In these experiments, the TrackBot switches between \textbf{Tracking mode} and the \textbf{Ranging mode} for ground truth estimation by switching its operating threads as well as the Openmote channel (since there is only one Openmote on the TrackBot). Step by step method of ranging are as follows.
\begin{enumerate}[leftmargin=*]
    \item Before ranging, the TrackBot and the LeaderBot finish up their last movement step and stops.
    \item TrackBot switches channel from 26 (Tracking channel) to 25 (Ranging Channel).
    \item TrackBot's Openmote sends a ranging request (REQ) packet to the Leader's RangeMote.
    \item Upon receipt of the REQ packet, the RangeMote and the LeaderBot prepares for ranging by temporarily switching off the remote control feature and sends a Ready (RDY) packet to the TrackBot.
    \item Upon receiving RDY packet, the TrackBot's Openmote turns ON the ultrasound-rf ping receiving mode by setting some flags in the MAC layer to prepare for interception of the packet and sends a GO packet.
    \item Upon receiving the GO packet,  the RangeMote on the 3pi sends exactly one RF packet and exactly one ultrasound ping @42kHz.
    \item If both transmissions are received, the TrackBot's Openmote estimates the TDoA and sends it to the mbed which then rotates the platform to the next orientation. If the TDoA process fails, the Openmote timeout and returns 0 to the mbed.
    \item After rotating the platform by one step, the mbed controls the Openmote to repeat the procedure from Step 3 to Step 7.
    \item If a full $360^\circ$ rotation of the platform is complete, the mbed processes the TDoA data to estimate the angle and the distance. The TrackBot's Openmote switches back to channel 26 for Tracking mode. 
\end{enumerate}

\begin{figure}[!ht] 
 \centering
    \subfloat[]{\label{fig:tdoa_performance_dist}\includegraphics[width=0.8\linewidth,height=0.5\linewidth]{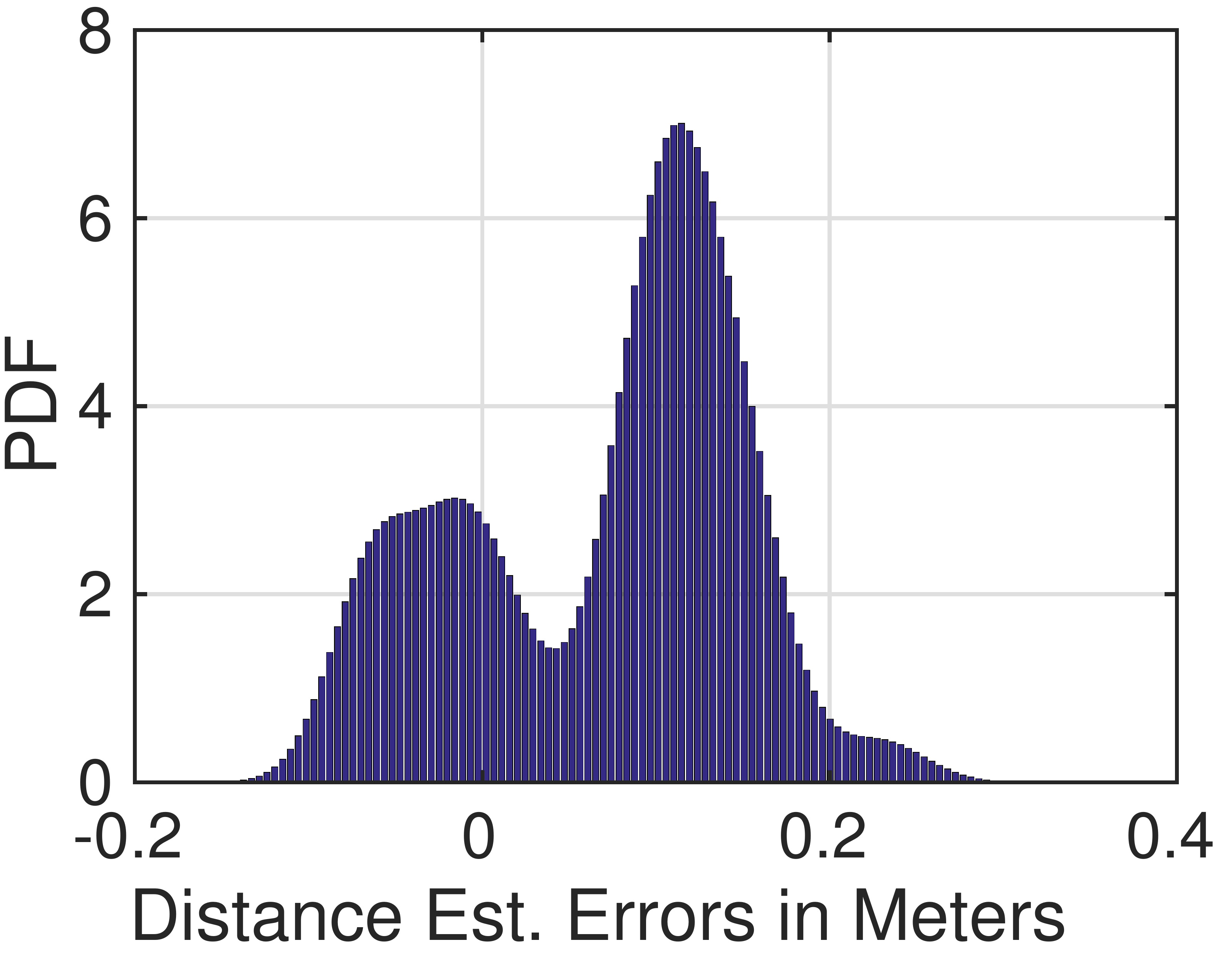}}\,
    \subfloat[]{\label{fig:tdoa_performance_angle}\includegraphics[width=0.8\linewidth,height=0.5\linewidth]{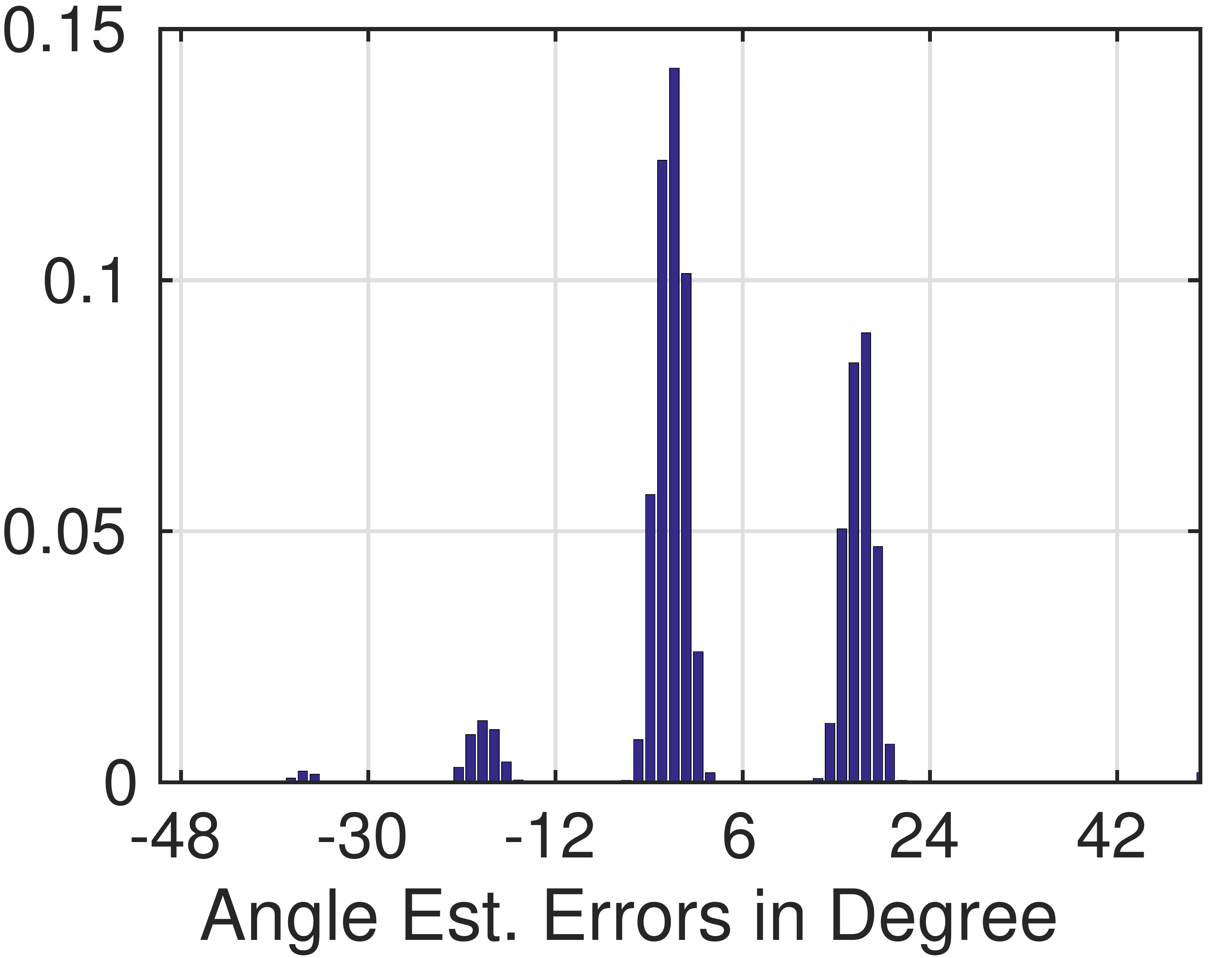}}
 \caption{ TDOA based Localization System Performance: (a) Distance Estimation Errors and (b) Angle Estimation Errors}
 \label{fig:tdoa_performance}
\end{figure}

Before evaluating the ARREST system on the basis of the TDoA ground truth system, we first evaluate the performance of the TDoA system. We found that the worst case distance estimation errors in TDoA systems are in the order of $10-20$ cm, as illustrated in Fig.~\ref{fig:tdoa_performance_dist}. The angle estimation statistics presented in Fig.~\ref{fig:tdoa_performance_angle}  demonstrates highly accurate performance in angle estimations. The slight chances of getting an error of $18^\circ$ is justifiable by our choice of ranging rotation step size of $18^\circ$. Thus, our TDOA system is accurate enough to be considered as a ground truth in line of sight situations. Nonetheless, we monitor the ranging outputs to trigger retries in case of very inaccurate outputs or momentary failures.
Moreover, in non-line sight situations, we still rely on manual measurements as the TDOA system fails in such scenarios.


\subsection{Different Experimental Settings}
With the aforementioned setup, we performed a range of experiments over months of duration with each run lasting for $1 - 2$ hours. For the ARREST setup, we use the Pragmatic policy with the weighted average angle estimation because of its superior performance in our emulations and small scale experiments. The LQG setup are also kept same as the small scale experiments. 
To diversify the situation we have performed experiments in four different classes of settings.


$\Box$ Large ($\geq 15m \times 10m$) office rooms with lots of computers, reflective surface, and cluttered regions. 

$\Box$ Long hallways ($\approx 200m$ long and $5-10m$ wide) with lots of turns.

$\Box$ Open ground floor spaces ($\approx 30m \times 30m$) with pillars.

$\Box$ Homelike environments with couches, furniture, and obstacles.

\subsection{Performance Analysis}
In Fig.~\ref{fig:practical_large_env_dist}, we present the statistics of the absolute distance between the TrackBot and the LeaderBot throughout the duration of the experiments in all four scenarios. Figure~\ref{fig:practical_large_env_dist} shows that the absolute distance is bounded by 3.5 meters in all four scenarios which further verifies our small scale experiment results presented in Fig.~\ref{fig:error_dist_practical}. Another noticeable fact from the figure is that ARREST system performs worst in the cluttered office scenarios which is justifiable due to presence of a lot of reflecting surfaces as well as obstacles. 

\begin{figure}[!ht] 
 \centering
 \includegraphics[width=\linewidth]{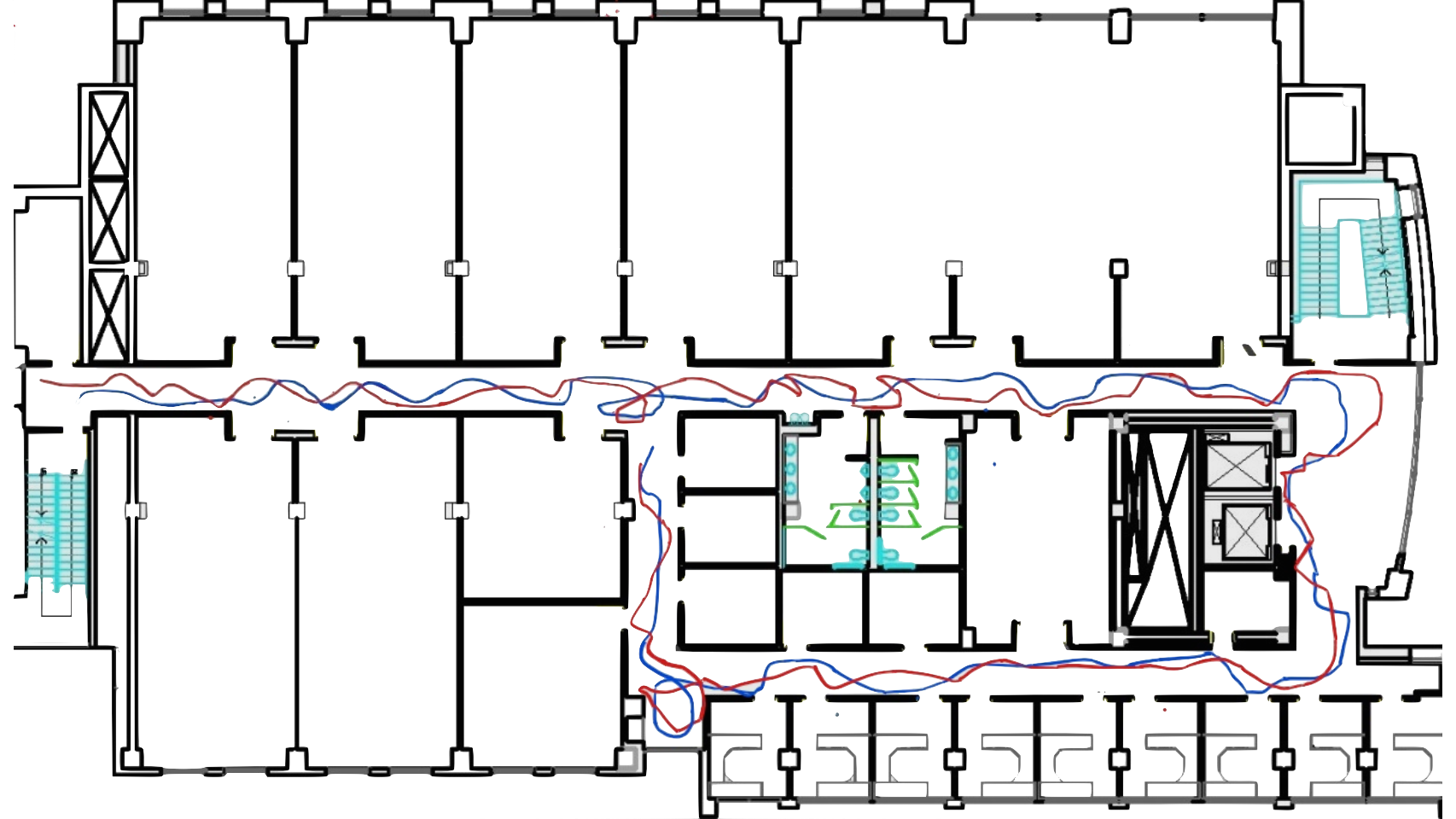}
 \caption{ Full path Trace for a Sample Large Scale Experiment (Blue$\implies$ Leader, Red$\implies$ TrackBot)}
  \label{fig:full_path_large_scale}
\end{figure}

Similar statistics can be seen in the absolute LQG distance error plot presented in the Fig.~\ref{fig:practical_large_env_dist_error}. Figure~\ref{fig:practical_large_env_dist_error} shows that the instantaneous absolute distance errors are $\leq 100cm$ with $\approx 90\%$ probability, except in the office scenario ($\approx 70\%$). {The comparatively higher distance errors for office scenarios is due to overestimation of distances in NLOS scenarios and in presence of strong multipath signals. However, this does not affect the performance much as the temporarily predicted higher distance tends to only lead to a temporary higher velocity of the TrackBot.} In summary, the distance error statistics is also mostly similar to the distance error statistics from the small scale experiments. Similar pattern can be observed in the angle estimation error plots presented in Fig.~\ref{fig:practical_large_env_angle_error}. Again the office space performance is worst. The open space performance is prominently better than the other scenarios due to absence of any sort of multipath signals. The instantaneous angle errors are less than $40^\circ$  with high probability ($\approx 85\%$) in \textbf{overall statistics.} However the scenario specific errors statistics (error being less that $40^\circ$) vary from $\approx 75\%$ probability in indoor setting to $\approx 100\%$ probability in the outdoor settings. {This slight discrepancy between small scale and large scale angle error performance is mainly due to different environment settings as evident from the Fig.~\ref{fig:practical_large_env_angle_error} itself.
In Fig.~\ref{fig:full_path_large_scale}, we present a sample illustrative trace of a large scale hallway experiment, drawn based on manual reconstruction from a video recording and markings on the floor.}

\subsection{Multipath Adaptation}
\label{sec:multipath_adapt}
Similar to small-scale experiments, we perform a set of experiments with a static Leader not in the line of sight of the TrackBot for $\geq 50\%$ of the TrackBot's path. Due to the TrackBot's ability to leverage a good multipath signal, the TrackBot was able to find the Leader in $70\%$ of the cases. 
However, we also notice that it fails dramatically if the TrackBot falls into a region with no direct path as well as no strong multipath signals (i.e., there exist multiple similar strength multipath signals). 
To overcome that, we add a \textbf{Multipath Angle Correction} module in the CAST layer (refer to Fig.~\ref{fig:arrest}). This module triggers a randomized movement for a single LQG period if: (1) the TrackBot hits an obstacle for $3-4$ consecutive LQG periods or, (2) the LQG estimated distance to the transmitter doesn't change much over $3-4$ consecutive periods. This policy basically leads the TrackBot to a random direction with the hope of getting out of such region. 
{However, we noticed that if the TrackBot keeps following randomized direction for consecutive LQG periods, it harms the tracking performance. Thus, we have set a minimum time duration (Five LQG periods in our implementation) between any two consecutive randomized movements. Note that, all these timing choices are made empirically via a range of real experiments. \emph{With this strategy, we noticed an improvement on the TrackBot's success rate from $\approx 70\%$ to a success rate of $\approx 95\%$ in such scenarios. However, the trade-off in such context is that the convergence in case of a far away Leader ($\geq 8m$) is now slower by $\approx 15 \%$. }
}

{\small
\begin{table}[!h]
    \centering
    \caption{Summary of Large Scale Real-World Experiments}
    \resizebox{\linewidth}{!}
    {\small
    \begin{tabular}{|m{\linewidth}|}
    \hline
    $\Box$ Absolute distance estimation errors are $<100cm$ with probability $\approx90\%$ except in the case of cluttered office environments.\\ 
    $\Box$ Average Absolute angle estimation errors are $<40^\circ$ with  probability $\approx 85\%$.\\
    $\Box$ The TrackBot stays within $3.5m$ of the Leader with probability $\approx100\%$ in all scenarios of tracking.\\
    $\Box$ In NLOS scenarios, addition of a conditional randomization improves the success rate from $70\%$ to $95\%$ but slows the converges by $\approx 15\%$ for static far-away Leader.\\
    \hline
    \end{tabular}
    }
    \end{table}
}

\section{Miscellaneous}
\subsection{Raw RSSI Data Analysis}
\label{sec: raw}
\begin{figure}[!ht] 
 \centering
 \subfloat[]{ \label{fig:raw_data_large_env_dist} \includegraphics[width=0.8\linewidth,height=0.42\linewidth]{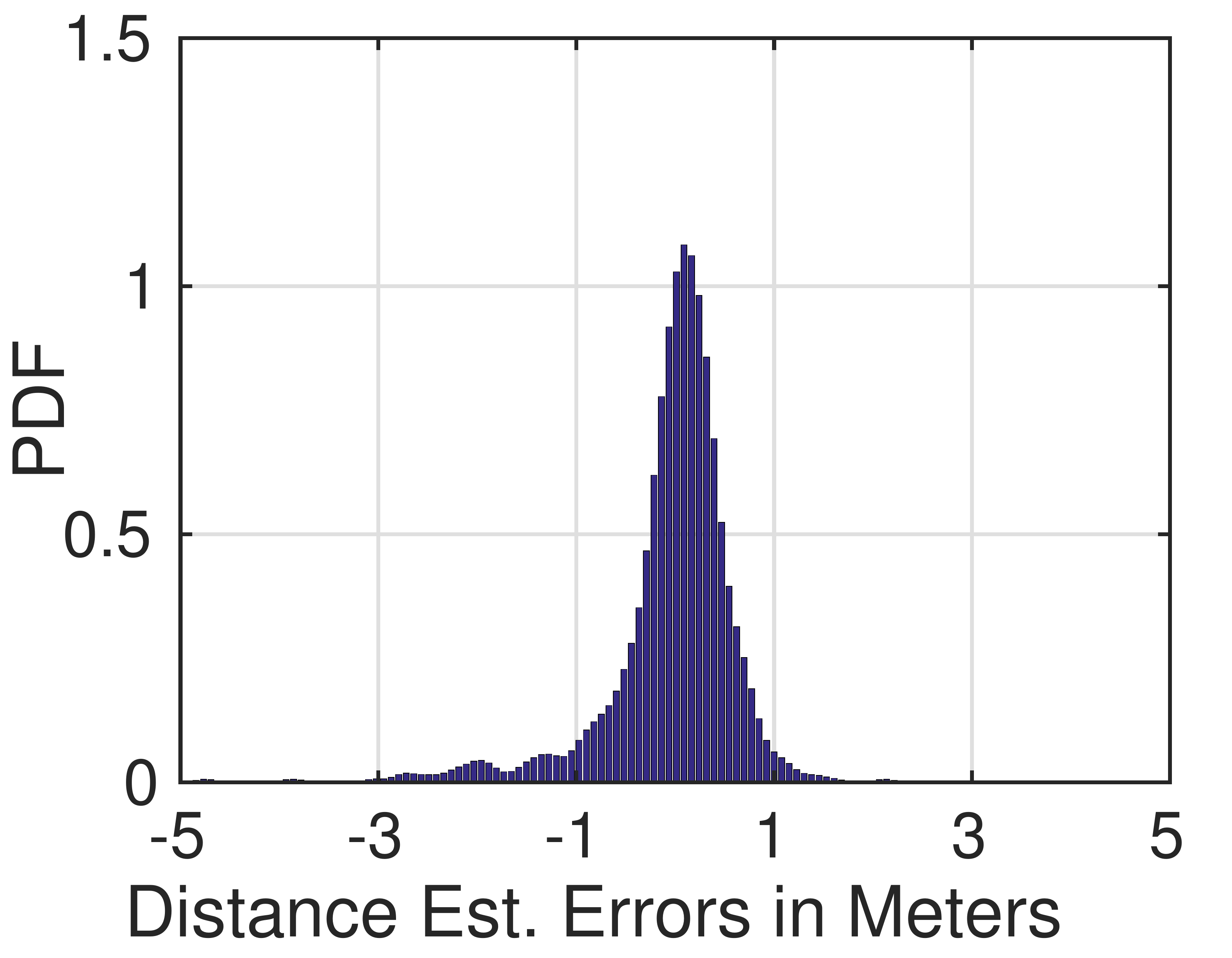}}\,
 \subfloat[]{ \label{fig:raw_data_large_env_angle}\includegraphics[width=0.8\linewidth,height=0.42\linewidth]{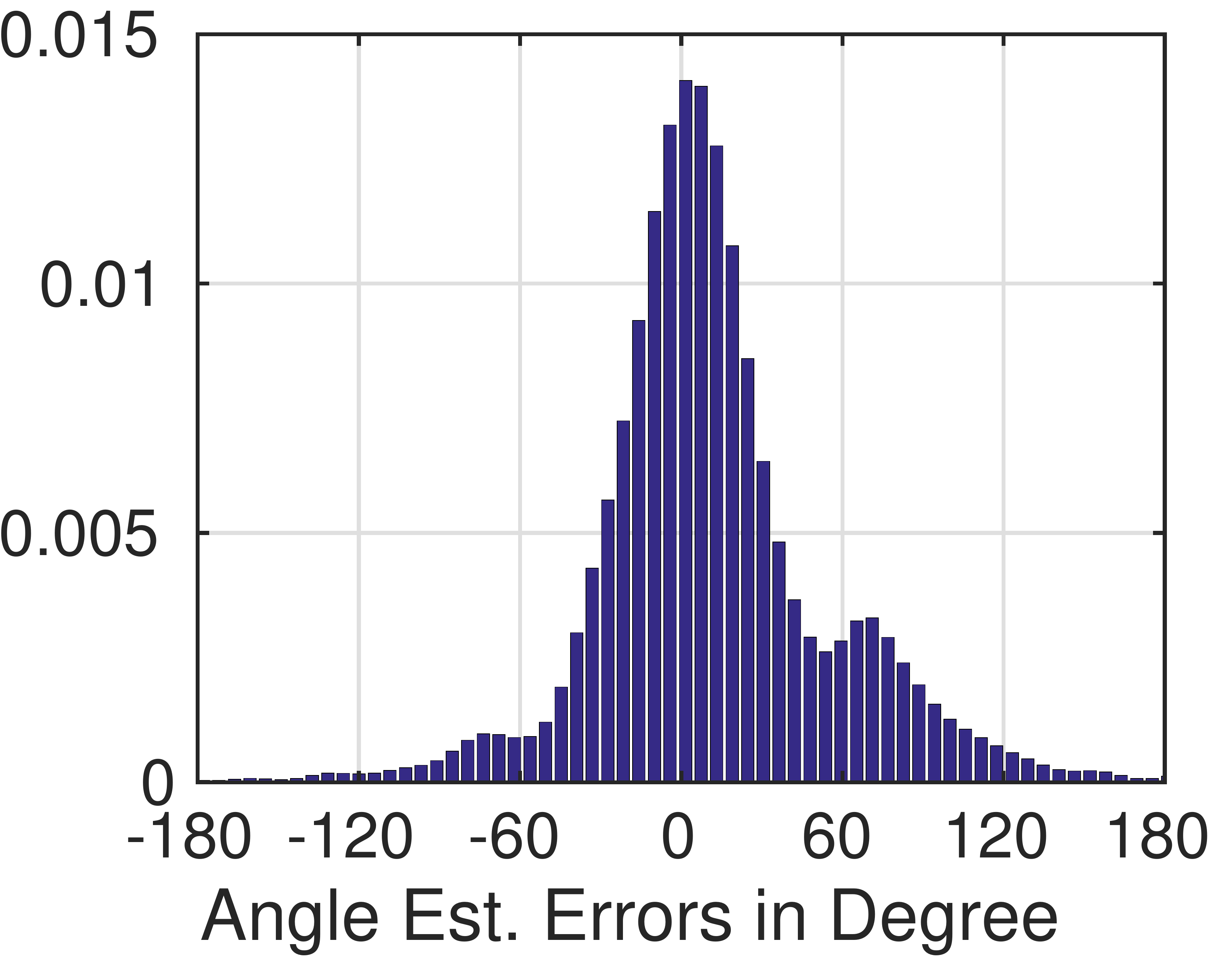}}
 \caption{Raw Data Analysis: (a) Distance Estimation Errors and (b) Angle Estimation Errors}
 \label{fig:raw_data_large_env}
\end{figure}

\begin{figure}[!ht] 
 \centering
 \subfloat[]{ \label{fig:error_stat_in_real_dist} \includegraphics[width=0.8\linewidth,height=0.42\linewidth]{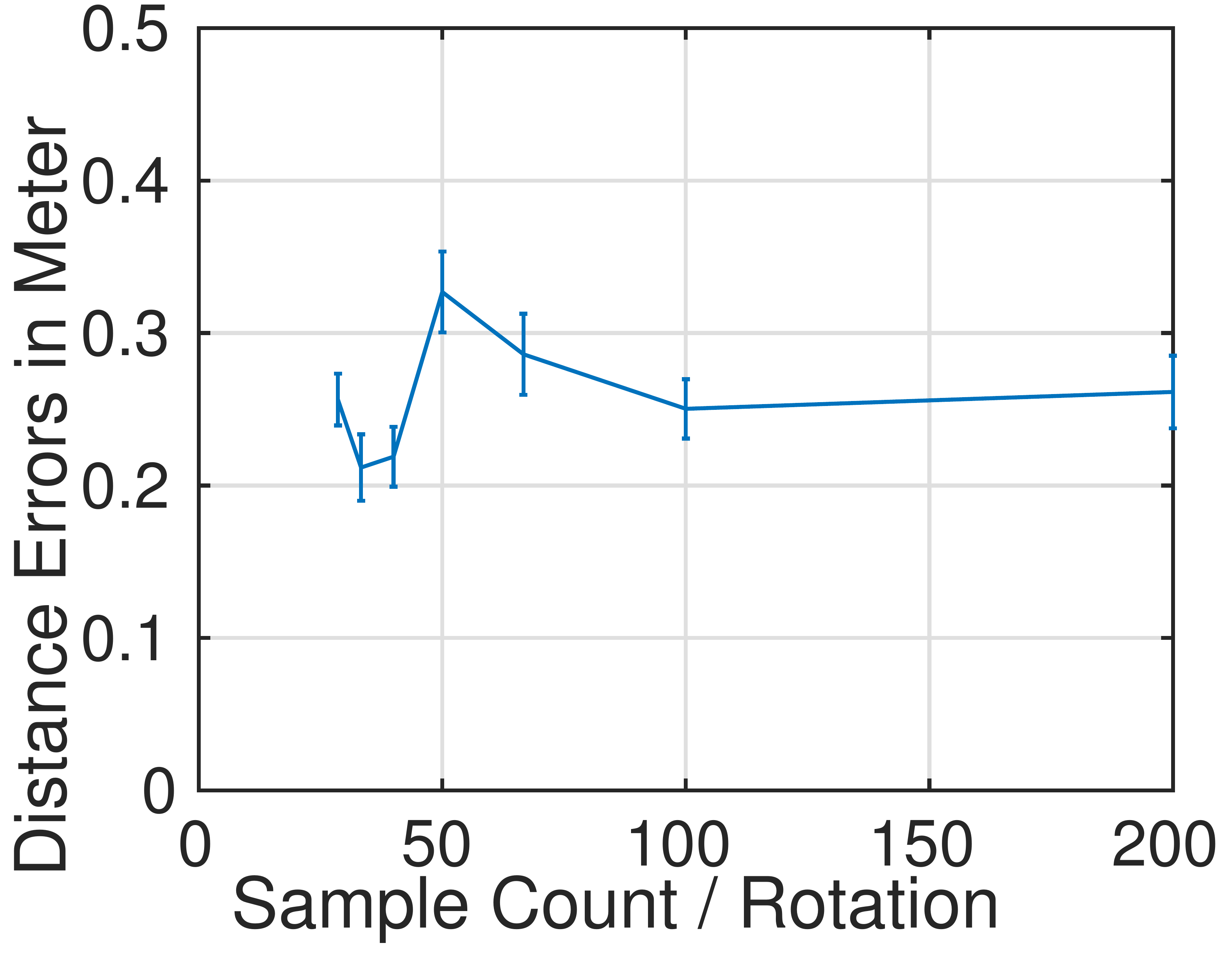}}\,
 \subfloat[]{ \label{fig:error_stat_in_real_angle}\includegraphics[width=0.8\linewidth,height=0.42\linewidth]{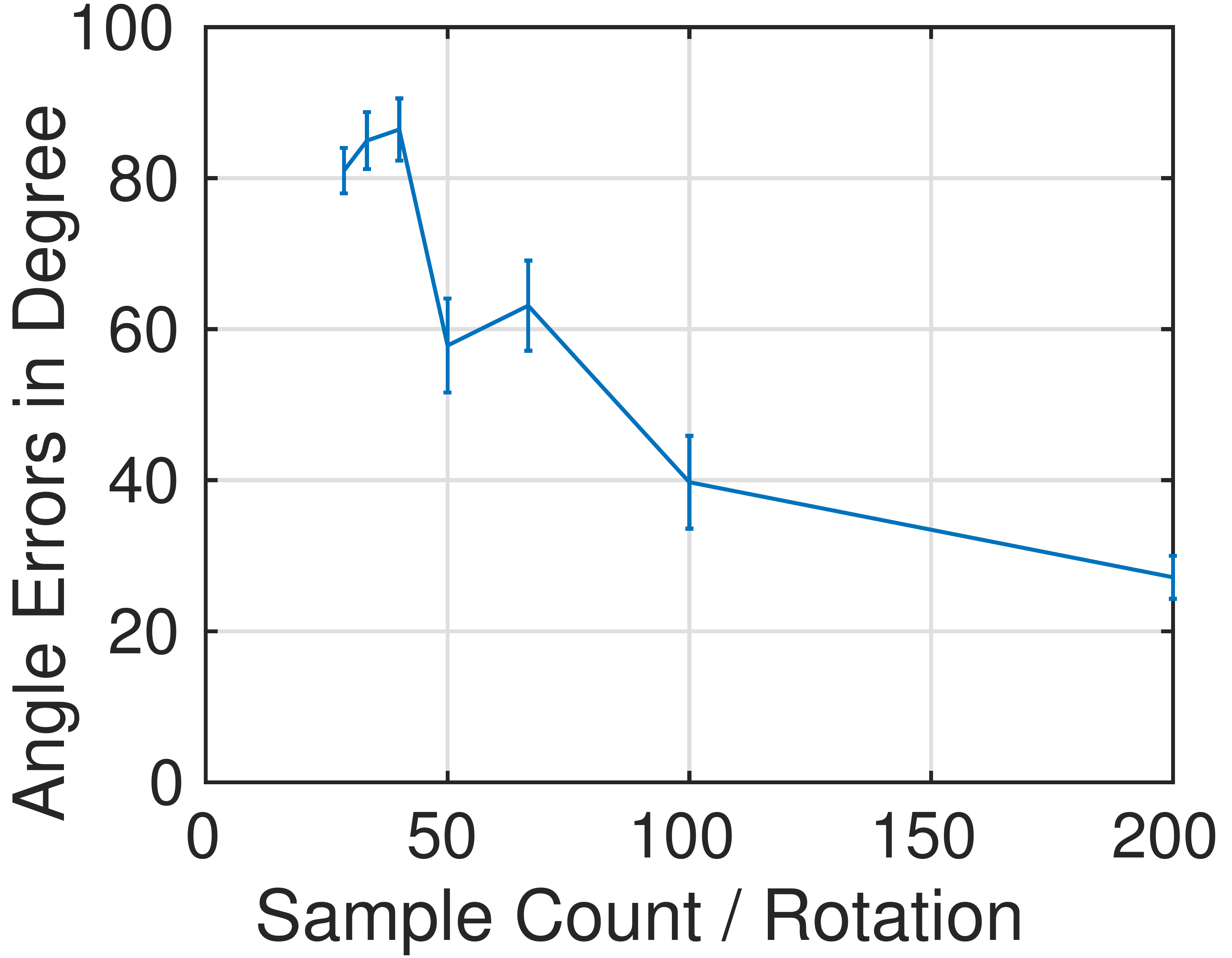}}
 \caption{Estimation Performance for Varying Sampling Rate: (a) Distance Estimation Errors and (b) Angle Estimation Errors}
 \label{fig:error_stat_in_real}
\end{figure}

Based on all our evaluations, we conclude that the presence of multipath signals does not hamper the performance if there exists a direct line of sight. 
To justify this further and to gather more insights on the systems performance, we perform a raw RSSI data analysis and calculate the unfiltered error statistics. 
In Fig.~\ref{fig:raw_data_large_env_dist}, we plot the RSSI pattern based distance estimation error statistics which demonstrates that the accuracy of the directional antenna pattern based distance estimations are in the order of less than 1 meter with $90\%$ probability. On the other hand, Fig.~\ref{fig:raw_data_large_env_angle} shows that the RSSI pattern based angle estimation error are less that $40^\circ$ with very high probability ($\approx 80\%$) with some deviations due to multipath and random changes in movement directions. Again, note that, the an error upto $40^\circ$ is acceptable due to our choice of directional antenna. {\emph{We also perform a set of experiments in an anechoic chamber with controlled position of the reflectors. While we do not present the respective plots for page limitations, the statistics are very similar to Fig.~\ref{fig:raw_data_large_env} for a maximum separation distance of 5m.} }
\todo{We also verify the performance of the RSSI based estimation for varying sampling rate. For these set of experiments, we fix the distance and angle between the TrackBot and the Leader and properly set the channel parameters before each experiment.} Figure~\ref{fig:error_stat_in_real} presents the average angle errors and average distance estimation errors with $95\%$ confidence interval for varying sampling rate. Figure~\ref{fig:error_stat_in_real_angle} shows that the angle estimation performance deteriorates as the sampling rate is decreased which is self-justified. The distance estimation actually doesn't vary much with the sampling rate. 


Our numbers may even appear to be better than those typically reported for RSSI based localization (where typical accuracies are $\approx 2m-5m$ or higher), but this is attributed to the fact that the distance estimates use the average of $40 - 200$ samples, one from each sample's respective antenna orientation. This analysis also suggests that we can use sampling rate of $100$ samples/rev to achieve similar performance. Nonetheless, we stick with $200$ sample/rev as we notice a loss of maximum $70-90$ samples per revolution in severe scenarios.


\subsection{Different Sensing Modalities}
\label{sec:modality}
\begin{figure*}[!ht] 
 \centering
   \subfloat[]{\label{fig:control_error_dist} \includegraphics[width=0.35\linewidth,height=0.20\linewidth]{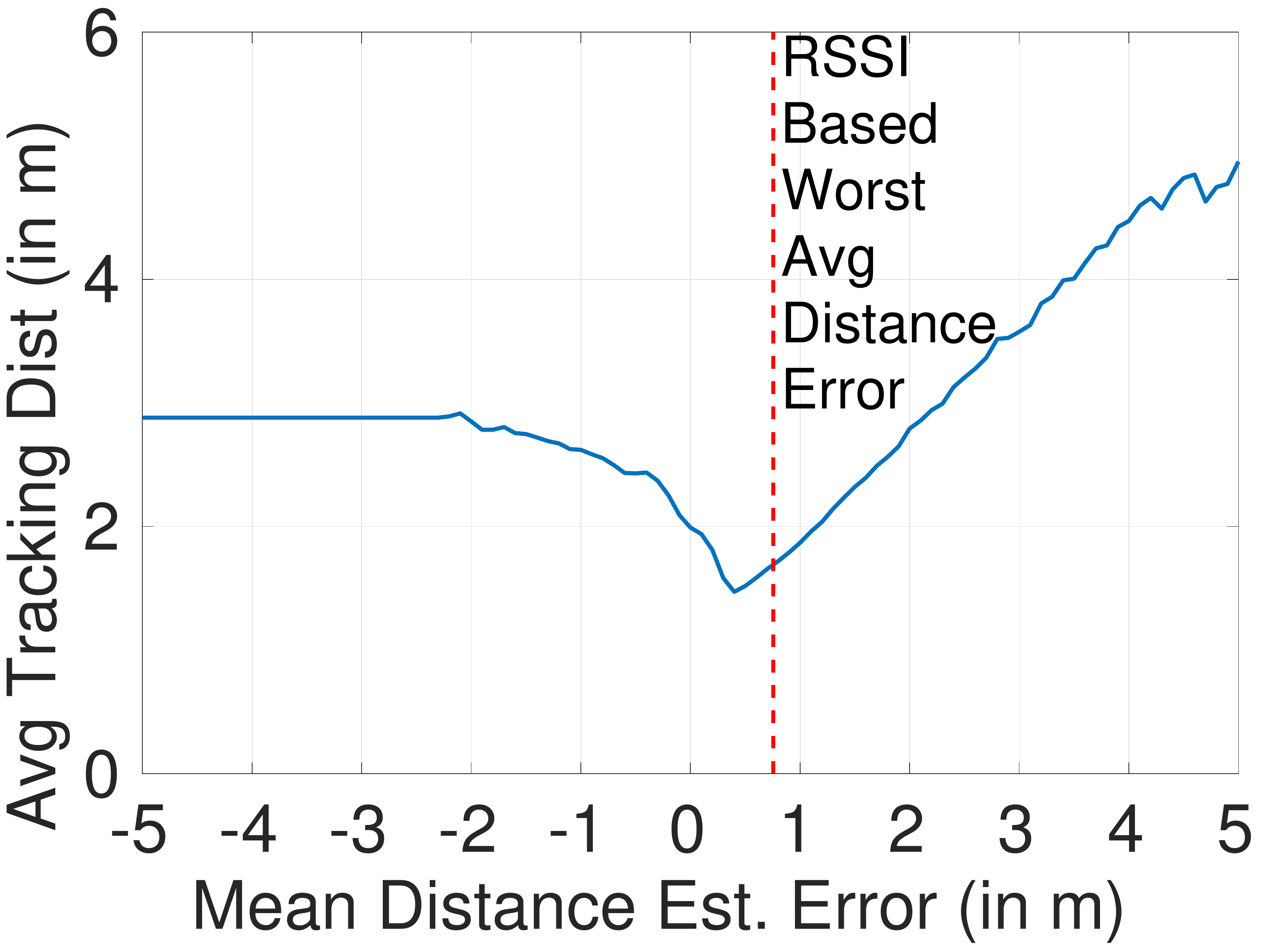}} \qquad
   \subfloat[]{\label{fig:control_error_angle}\includegraphics[width=0.35\linewidth,height=0.20\linewidth]{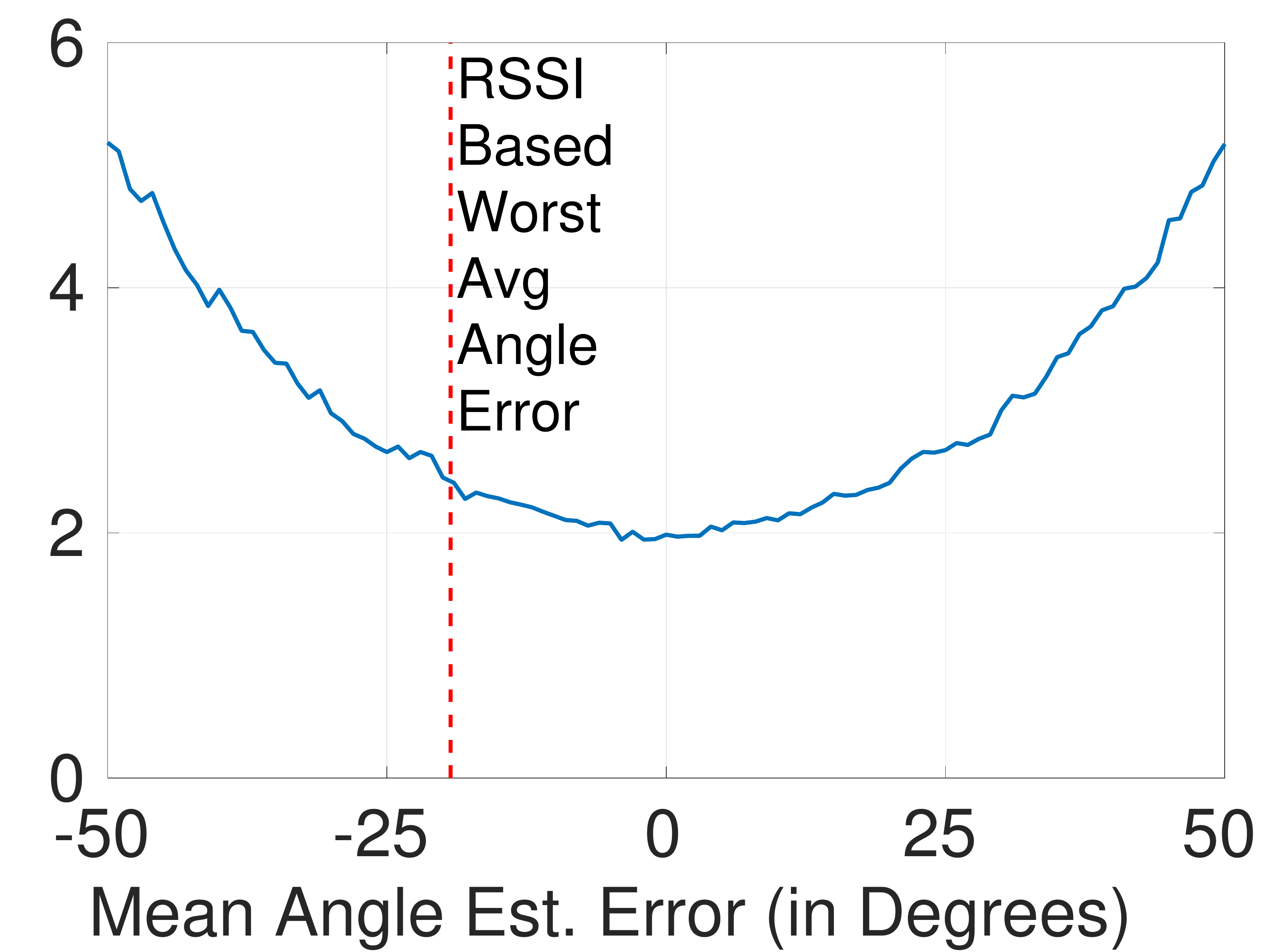}}
 \caption{ Performance of the ARREST System in Terms of Controlled Estimation Errors}
 \label{fig:control_error}
\end{figure*}

While our proposed ARREST architecture employs pure RSSI based distance, angle, and speed estimations, the same architecture can be easily adapted to use other technologies such as cameras or infrared sensors. 
In such cases, we just need to modify the CANE layer of the ARREST architecture and feed the relative position approximations to the CAST layer. 
Now, each of these estimation technologies i.e., camera based or RF based estimations, have different accuracies in terms of distance and angle estimations. 
To analyze the tracking performance of the ARREST system, oblivious to the actual technology used in CANE layer, we perform a set of simulation experiments where we control the average errors in the distance and the angle estimations.
Figure~\ref{fig:control_error_dist} illustrates one instance of such experiments where we fix the average angle error ($0$ in this case)
and vary the average distance estimation error.
{Figure~\ref{fig:control_error_dist} shows that the effect of positive estimation errors ($d_{org}-d^e>0$, where $d_{org}$ is the actual distance)
have a more detrimental effect on the tracking performance than negative errors.}

This is justified as positive distance estimation errors imply always falling
short in the movements, whereas, negative errors imply over-estimations and more
aggressive movements. It is also noticeable that there exists an optimal value
of average distance estimation error. The value of this optimal distance error
depends on the maximum Leader speed as well as average angle error. Next, we
plot the relation between average tracking distance and average angle error
while the average distance error is kept to be $0$ in
Fig.~\ref{fig:control_error_angle}. It is obvious and quite intuitive that the
best tracking performance is obtained for an average angle estimation error of
$0$. Note that, we do not control the speed error separately as it is directly
related to the angle and distance estimations.  This analysis demonstrates the
versatility of our ARREST architecture to tolerate a large range of estimation
errors. More specifically, it tolerates up to $5m$ average distance error and
$45^\circ$ average absolute angle error in a successful tracking application.
This analysis also shows that while RSSI based system is not optimum, it has
reasonable performance compared to the best possible ARREST system (with zero
distance and angle estimation error).

\subsection{Some Challenges and Lesson Learned}
\label{sec:lesson}

Here we present two main challenges we faced in this project along with our
methodology to overcome them.

\emph{Rotating Platform Wire Twisting:} 
To overcome this challenge mechanically, we alternate the rotation direction between clockwise and anticlockwise. Further, we opt for a system design where every device on the rotating platform (in our case only ultrasound) communicate via the serial line between the Openmote (on the platform) and the mbed. To achieve that we use multi-threading in RIOT OS and the HDLC protocol to allocate a dedicated thread and HDLC identifier for each of the peripheral device on the Openmote. Upon receipt of a HDLC packet from the Openmote, the mbed process the HDLC identifier to identify the source device and perform the necessary operation. 

\emph{Missing Samples:} During the RSSI sampling, we noticed that the mbed receives a very low number of samples from the Openmote with chunks of missing samples. This was caused by the beaconer's buffer overflow (due to continuous beaconing), interference from other devices, and loss in  the communication between the Openmote and the mbed. To solve the beaconer buffer overflow issue, we added a periodic reset controller on the 3pi LeaderBot's mbed that resets the beaconer after every two minute via a GPIO pin. The interference and noise related missing samples problem were solved by changing the beaconing from broadcast to unicast and also via employing our proposed block based angle estimation (Refer to Section~\ref{sec:arrest_angle}). We reduce the loss due to communication between the mbed and the Openmote by employing the HDLC protocol based packetized serial communication and proper ACK mechanism. 


\section{Conclusion}

While our proposed solely RSSI based relative localization and tracking system for autonomously following a RF-emitting object works with reasonable performance, 
there are a lot of research questions that need to be addressed in our future works and are not part of this work. First, we intend to develop a strategy with a proper trade-off between Optimism and Pragmatism, which will potentially improve the performance. Second, we want to make the system faster by employing the concept of compressive sampling that will potentially allow for continuous-time decision making. Third, we want to explore the optimal configuration options for our system as well as the optimality conditions for RF based tracking. Fourth, we intend to look into more structured randomization in the TrackBot's movements to improve performance in severe NLOS environments. Finally, we intend to explore the domains of game theory and robust control to see if better or more robust predictions of the Leader's motion could improve the performance. 


{
\bibliographystyle{IEEEtran}
\bibliography{ref}
}

\end{document}